\pdfoutput=1
\documentclass{article}

\PassOptionsToPackage{numbers,compress}{natbib}

\usepackage[main, final]{neurips_2026}

\usepackage[utf8]{inputenc} %
\usepackage[T1]{fontenc}    %
\usepackage{hyperref}
\usepackage{url}            %
\usepackage{booktabs}       %
\usepackage{multirow}       %
\usepackage{amsmath}
\usepackage{amssymb}
\usepackage{amsfonts}       %
\usepackage{graphicx}
\usepackage{subcaption}
\usepackage[capitalise,noabbrev]{cleveref}
\usepackage{nicefrac}       %
\usepackage{microtype}      %
\usepackage[dvipsnames]{xcolor}         %
\usepackage[colorinlistoftodos]{todonotes}
\usepackage{enumitem}
\usepackage{xspace}
\usepackage{textcomp}
\usepackage{wrapfig}

\hypersetup{colorlinks=true,linkcolor=black,citecolor=blue!50!black,urlcolor=blue!50!black}

\usepackage[dvipsnames]{xcolor}
\usepackage{soul}
\usepackage{ifthen}

\definecolor{MacroColorMath}{RGB}{0,0,0}

\newcommand{\myMathMacro}[2]{%
  \newcommand{#1}{\textcolor{MacroColorMath}{\ensuremath{#2}}}%
}

\myMathMacro{\xseq}{\mathbf{x}}              %
\myMathMacro{\hvec}{\mathbf{h}}              %
\myMathMacro{\wvec}{\mathbf{w}}              %
\myMathMacro{\bvec}{\mathbf{b}}              %

\myMathMacro{\Wmat}{\mathbf{W}}              %
\myMathMacro{\Umat}{\mathbf{U}}              %
\myMathMacro{\Util}{\widetilde{\mathbf{U}}}  %
\myMathMacro{\Gmat}{\mathbf{G}}              %
\myMathMacro{\Pmat}{\mathbf{P}}              %
\myMathMacro{\Qmat}{\mathbf{Q}}              %
\myMathMacro{\Imat}{\mathbf{I}}              %

\myMathMacro{\gvec}{\mathbf{g}}              %

\definecolor{MacroColorText}{RGB}{0,0,0}

\newcommand{\myTaskMacro}[2]{%
  \newcommand{#1}{\textcolor{MacroColorText}{\texttt{#2}}\xspace}%
}

\myTaskMacro{\unigramlp}{UnigramLogProb}
\myTaskMacro{\bigramlp}{BigramLogProb}
\myTaskMacro{\trigramlp}{TrigramLogProb}
\myTaskMacro{\bigrampmi}{BigramPMI}
\myTaskMacro{\trigrampmi}{TrigramPMI}
\myTaskMacro{\pos}{POS}
\myTaskMacro{\deplen}{DependencyLength}
\myTaskMacro{\deprel}{DependencyRelation}
\myTaskMacro{\depdepth}{DependencyDepth}
\myTaskMacro{\ner}{NER}
\myTaskMacro{\ssnoun}{SupersenseNoun}
\myTaskMacro{\ssverb}{SupersenseVerb}

\definecolor{MacroColorMetric}{RGB}{0,0,0}

\newcommand{\myMetricTextMacro}[2]{%
  \newcommand{#1}{\textcolor{MacroColorMetric}{#2}\xspace}%
}

\newcommand{\myMetricMathMacro}[2]{%
  \newcommand{#1}{\textcolor{MacroColorMetric}{\ensuremath{#2}}}%
}

\myMetricTextMacro{\cetext}{Concept Energy}
\myMetricMathMacro{\ce}{\mathrm{CE}}

\myMetricTextMacro{\oetext}{Optimization Engagement}
\myMetricMathMacro{\oem}{\mathrm{OE}}

\definecolor{MacroColorOperator}{RGB}{0,0,0}

\newcommand{\myOperatorMacro}[2]{%
  \newcommand{#1}{\textcolor{MacroColorOperator}{\ensuremath{#2}}}%
}

\myOperatorMacro{\PMI}{\mathrm{PMI}}
\myOperatorMacro{\tr}{\operatorname{tr}}
\myOperatorMacro{\rk}{\operatorname{rank}}

\newif\ifshowtodos
\showtodosfalse   %

\ifshowtodos
    \newcommand{\mytodo}[1][]{%
        \sethlcolor{lime}
        \ifthenelse{\equal{#1}{}}{%
            \hl{TODO}}{%
            \hl{TODO: #1}%
        }
    }
    \newcommand{\tocite}[1][]{%
        \sethlcolor{yellow}
        \ifthenelse{\equal{#1}{}}{%
            \hl{CITE}}{%
            \hl{CITE: #1}%
        }
    }
    \newcommand{\alex}[1][]{%
        \sethlcolor{SkyBlue}
        \ifthenelse{\equal{#1}{}}{%
            \hl{ALEX}}{%
            \hl{ALEX: #1}%
        }
    }
    \newcommand{\yupei}[1][]{%
        \sethlcolor{pink}
        \ifthenelse{\equal{#1}{}}{%
            \hl{YUPEI}}{%
            \hl{YUPEI: #1}%
        }
    }
\else
  \newcommand{\mytodo}[1][]{}
  \newcommand{\tocite}[1][]{}
  \newcommand{\alex}[1][]{}
  \newcommand{\yupei}[1][]{}
\fi

\title{Geometrically Disentangling Concept Learning\\from the Language Modeling Loss}

\author{%
  Yupei Wang$^{1}$ \quad
  Neil Mallinar$^{1}$ \quad
  Mikhail Belkin$^{1}$ \quad
  Alex Warstadt$^{1,2}$ \\
  $^{1}$Hal{\i}c{\i}o\u{g}lu Data Science Institute \quad
  $^{2}$Department of Linguistics \\
  University of California San Diego \\
  \texttt{\{yuw306,nmallina,mbelkin,awarstadt\}@ucsd.edu}
}

\begin{document}

\maketitle

\begin{abstract}
    Language models acquire a vast array of concepts and abilities from next-token prediction alone.
    We introduce an interpretability framework that exposes how this single loss relates to the development of different concepts at different points during pretraining. 
    We propose \cetext, a measure of the concept-specific alignment between a concept subspace and the average gradient outer product (AGOP) of the next-token-prediction loss. 
    The concept subspace captures where the concept currently lives in the model's representation; AGOP captures the directions to which the loss is most sensitive.
    Across 19 checkpoints of Pythia-410M and -1B and 12 lexical, syntactic, and semantic concepts, we find that the first peak in \cetext generally occurs at or before the steepest increase in probe decodability.
    Furthermore, whereas probe performance usually increases monotonically throughout training, \cetext is not confined to a single peak; some concepts exhibit sustained or repeated engagement across pretraining, suggesting that concept learning may require continual reinforcement or occur at multiple points.
    Our results indicate that the loss contributes to concept learning in a myriad of ways that simple probe performance alone cannot reveal.\footnote{%
    The project repository is available at \url{https://github.com/YpLarryWang/concept-energy}.
    }
\end{abstract}

\section{Introduction}
\label{sec:intro}

Large language models (LLMs) acquire different concepts and capabilities throughout pretraining, but the aggregated training loss cannot reveal when and how particular behaviors, circuits, and features appear, stabilize, or reorganize over the course of training.
A growing body of work examining behavioral learning curves, e.g., evaluating on downstream tasks, has found that LLMs seem to abruptly acquire some abilities.
Emergence is often defined as the abrupt appearance of abilities as models scale~\citep{wei2022emergent}; 
grokking is the training-time counterpart, in which test accuracy jumps long after the training set has been memorized~\citep{power2022grokking}.
However, behavioral measures can make model development appear more abrupt than measures of internal changes.
In several controlled settings, such transitions are preceded by gradual internal reorganization that is invisible to standard evaluation metrics:
hidden progress in sparse-parity learning~\citep{barak2022hidden}, circuit formation before the grokking transition in modular arithmetic~\citep{nanda2023progress,pmlr-v267-mallinar25a}, and early residual-stream restructuring before in-context learning improves~\citep{olsson2022context,hoogland2025loss-landscape}.
These results motivate a \emph{geometric-precedence hypothesis}: concept-specific signals in the loss geometry precede the largest improvements in concept decodability or model behavior. 
Uncovering these hidden dynamics requires studying full training trajectories rather than isolated final-checkpoint snapshots, a paradigm known as developmental interpretability~\citep{sharkey2025openproblems}.

Existing work approaches developmental interpretability from two complementary starting points: \emph{concept-centric} and \emph{optimization-centric}.
\emph{Concept-centric} analyses start from a concept of interest, define a proxy (e.g., a probe or a downstream evaluation) or representation, and track the changes of the performance of the proxy or the representation directly across training checkpoints~\citep{chiang-etal-2020-pretrained,liu-etal-2021-probing-across,saphra-lopez-2019-understanding,muller-eberstein-etal-2023-subspace,arora-etal-2024-causalgym}.

\emph{Optimization-centric} approaches instead start from a training objective or its artifacts, like its loss landscape, its gradients or an auxiliary reconstruction objective, and then use these signals to identify interpretable staged patterns of concept dynamics throughout training~\citep{hoogland2025loss-landscape,wang2025differentiation,ge2026evolution,bayazit-etal-2026-crosscoding}.

In this work, we bridge the \emph{concept-centric} and \emph{optimization-centric} views. 
We begin with a set of predefined concepts of interest (e.g., part-of-speech tag, named entity type), operationalized as low-rank linear subspaces induced by probes.
We then ask, \emph{when during pretraining does the loss geometry actively concentrate on a concept subspace?}
To address this, we examine the loss geometry within concept subspaces.
Specifically, we define a metric called \cetext to measure how much of next-token loss sensitivity is concentrated on a particular concept at a certain point during pretraining.
\cetext is based on the projection of the average gradient outer product (AGOP; \citealp{radhakrishnan2024mechanism,boix-adsera2026fact}), which identifies the directions in representation space to which the next-token loss is most sensitive, onto the concept subspace.
A probe tells us whether a concept can be linearly decoded from the representations; \cetext tells us whether the next-token loss gradients actively concentrate along the directions that encode it.

This perspective motivates two research questions about geometric precedence:
\begin{itemize}[leftmargin=2em,topsep=2pt,itemsep=2pt]
    \item \textbf{RQ1: Decodability.} Does the geometric transition revealed by \cetext precede the largest observed improvement in concept decodability?
    \item \textbf{RQ2: Behavior.} Does the same order hold for probe-independent model behavior?
\end{itemize}

\begin{figure}[t!]
    \centering
    \includegraphics{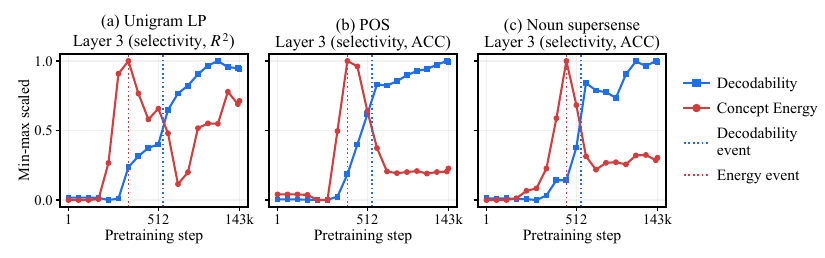}
    \setlength{\abovecaptionskip}{3pt}
    \caption{
        Examples of decodability and \cetext changes throughout Pythia-1B pretraining.
        All examples are taken from the 3rd layer.
        The \unigramlp example shows an early local \cetext peak followed by renewed engagement, ending high.
        The \pos and \ssnoun examples instead show one clear \cetext peak followed by lower levels.
        Decodability is probe selectivity (\Cref{subsec:concept-subspaces}).
    }
    \label{fig:example-decodability-vs-energy}
\end{figure}

We find evidence for geometric precedence in both Pythia-410M and -1B~\citep{biderman2023pythia}.
The earliest clear \cetext event precedes the steepest rise in decodability in 11 and 9 of the 12 probing tasks, respectively.
\cetext provides a distinct perspective on top of probe performance alone; several concepts whose probe performance rises steadily and consolidates at a plateau nevertheless exhibit \cetext curves that rise, form a peak and then go down, and even have multiple distinct peaks across pretraining (see examples in \Cref{fig:example-decodability-vs-energy}).
This suggests that the next-token loss can exert pressure on a given concept subspace in a multi-phase pattern.
We further investigate the ordering between \cetext and behavior by comparing a subject--verb agreement probe with the model's grammaticality performance on the minimal-pair benchmark BLiMP~\citep{warstadt-etal-2020-blimp-benchmark}, and the divergence between models' output distribution and reference n-gram distribution.
In both cases, the energy event precedes the behavioral event.
Together, these results suggest that geometric precedence is a common pattern across tasks, and that the loss geometry within a concept subspace, concept decodability, and model behavior can follow distinct trajectories.

\section{Related Work}
\label{sec:rw}

\subsection{Conceptual representations inside neural language models}

A significant thread of interpretability research aims to decode features from the internals of neural LMs.
One of the simplest and most widely-used methods to extract concept representations from LM internals is linear probing. 
Probes measure whether a concept is linearly decodable from the model internals and provide a simple linear readout of that concept's representation~\citep{ettinger-etal-2016-probing,adi2017fine-grained,belinkov-etal-2017-neural,belinkov-etal-2017-evaluating,hewitt-manning-2019-structural}.
A probe can reach high performance by memorizing word identities; control tasks, which assign random labels to word types, were therefore introduced to measure concept selectivity of a probe~\citep{hewitt-liang-2019-designing}.

However, decodability does not imply that the model relies on these features during prediction. 
\citet{elazar-etal-2021-amnesic} located the decodable direction of a feature of interest in the representation space using a linear probe, then estimated the causal contribution of that information to model behavior by measuring the performance change after removing those directions.
A broader line of work seeks representations that can causally influence model behavior through intervention, going beyond probes altogether~\citep{vig-etal-2020-causal-mediation-analysis,marks-2024-the-geometry-of-truth,arora-etal-2024-causalgym,arditi2024refusal}.
These efforts provide empirical grounding for the \textit{linear representation hypothesis}~\citep{park-2023-the-linear-repr-hype,sharkey2025openproblems}, which holds that concepts form linear subspaces in neural model representations.

\subsection{The development of concept representation during training}

Following the organization introduced in \Cref{sec:intro}, we distinguish two lines of work. 

\textbf{Concept-centric}\quad
\citet{chiang-etal-2020-pretrained} and \citet{liu-etal-2021-probing-across} measured the decodability of specific linguistic concepts using proxy models such as probes across training checkpoints, tracking when the model can be said to have ``\emph{acquired}'' each concept.
However, this approach rests on an implicit assumption: that changes in probe performance across training directly reflect the model's acquisition process during pretraining. 
This is a claim that has \textit{not} been independently validated.
A more precise approach moves beyond proxy performance and instead observes changes in the geometry of internal representations directly. \citet{saphra-lopez-2019-understanding} studied learning dynamics by comparing representations across time between a next-token-prediction language model and an independently trained next-tag prediction model.
\citet{muller-eberstein-etal-2023-subspace} measured principal angles between probe-defined task-specific subspaces at different training checkpoints to track how these task-specific subspaces shift and interact over the course of training.

\citet{arora-etal-2024-causalgym} moved beyond probes to causally link representations to model behavior.
They found linear intervention directions via Distributed Alignment Search across layers and token positions, and used the resulting causal effects on targeted syntactic structures to study the stagewise emergence of linguistic mechanisms.
There are also works that directly address the development of output behavior.
\citet{chang-bergen-2022-word} and \citet{chang-etal-2024-characterizing} investigated the development of language models' next-token predictions.
They found that model predictions initially align with and then diverge from the unigram distribution, before successively approaching and diverging from higher-order n-gram distributions.

\textbf{Optimization-centric}\quad
\citet{hoogland2025loss-landscape} showed that the local learning coefficient (LLC), a measure of loss-landscape geometry, can reveal the developmental stages of the overall model.
\citet{wang2025differentiation} further showed that the refined LLC can reveal the developmental stages of individual modules like self-attention. 
Methods based on sparse dictionary learning also fall into this category, since proxies such as sparse autoencoders are not conditioned on any known concept during training. By training crosscoders~\citep{lindsey2024sparse-crosscoders} shared across checkpoints, \citet{ge2026evolution} and \citet{bayazit-etal-2026-crosscoding} used concepts discovered via sparse dictionaries to characterize the dynamics of conceptual structure throughout training.

\section{Aligning Concepts with Loss Geometry}
\label{sec:theoretical-method}

We introduce a metric we call \textbf{\oetext} to quantify the extent to which next-token loss gradients align with the subspace of a particular concept in the representation space, and further isolate the concept-specific component of this alignment via control tasks, which we call \textbf{\cetext}.
This framework applies to any specified linear concept subspace, regardless of how it is identified.
Specifically, we use subspaces identified by linear probes and quantify their alignment with next-token loss gradients using AGOP in this work.

\subsection{Probe-defined concept subspaces}
\label{subsec:concept-subspaces}

Let \(\hvec_{\ell,t}(\xseq,s)\in\mathbb{R}^d\) be the residual-stream activation at layer \(\ell\), checkpoint \(t\), sequence \(\xseq\), and token position \(s\).
For each task, layer, and checkpoint, we train a linear probe on these activations: a \emph{classification} probe is the softmax linear model \(p(y\mid \hvec)=\operatorname{softmax}\left(\Wmat(\hvec-\bar{\hvec})+\bvec\right)\) with weight matrix \(\Wmat\in\mathbb{R}^{C\times d}\), residual activation mean \(\bar{\hvec}\), and bias \(\bvec\); a \emph{regression} probe is the linear model \(\hat y = \wvec^{\top}(\hvec-\bar{\hvec})+b_0\) with weight vector \(\wvec\in\mathbb{R}^d\).
We center the probe inputs by subtracting the mean activation $\bar{\hvec}$ of the training set, following \citet{burns2023discovering,marks-2024-the-geometry-of-truth}.\footnote{%
    We explain the rationale for mean-centering and compare results with and without mean-centering in \Cref{app:centring}.
}
For every probe, we also train a probe on a control task with the same inputs.
We operationalize linear decodability as selectivity~\citep{hewitt-liang-2019-designing}: the held-out score of the probe (classification accuracy or regression \(R^2\)) on the real task minus that of a probe trained on that control task, where labels are randomly assigned to input types (e.g., word types).
Selectivity thus discounts the part of the score that a probe can obtain by memorizing a label for each input type, isolating the decodability driven by the concept itself. 
We compare the results of measuring decodability using selectivity versus raw probe scores in \cref{app:selectivity}.

We define the concept subspace \(\Umat_{\ell,t}\) from the trained probe weights.
For a \textit{classification} probe, we subtract the row mean from \(\Wmat\) and take its leading \(C-1\) right singular vectors, where \(C\) is the number of classes used to train the probe.
For a \textit{regression} probe, \(\Umat_{\ell,t}\) is the normalized rank-one direction of the learned linear regressor.
These subspaces are operational definitions of concepts: they identify directions that support the extraction of concept information from the model representations.

\subsection{AGOP: loss sensitivity of next-token prediction}

To find directions for the next-token loss sensitivity in the representation space, we use the Average Gradient Outer Product (AGOP; \citealp{radhakrishnan2024mechanism,boix-adsera2026fact}) of the next-token cross-entropy loss.
Let \(\mathcal{X}\) be a set of evaluation sequences, and let \(\mathcal{S}_{\xseq} \subseteq \{1, \ldots, |\xseq|-1\}\) denote the selected next-token prediction positions in sequence \(\xseq\).
Let \(\theta_t\) denote the model parameters at training step \(t\), \(L_{\theta_t}(\xseq,s)\) be the next-token prediction loss at position $s$ of $\xseq$, and \(\gvec_{\ell,t}(\xseq,s)
= \nabla_{\hvec_{\ell,t}(\xseq,s)}
L_{\theta_t}(\xseq,s)\) be the loss gradient with respect to the activations at layer \(\ell\) and step \(t\).
Following \citet{radhakrishnan2024mechanism}, we define the layer-wise AGOP matrix as
\begin{equation}
  \Gmat_{\ell,t}
  =
  \frac{1}{N}
  \sum_{\xseq \in \mathcal{X}}
  \sum_{s \in \mathcal{S}_{\xseq}}
  \gvec_{\ell,t}(\xseq,s)\,
  \gvec_{\ell,t}(\xseq,s)^{\top},
  \qquad
  N = \sum_{\xseq \in \mathcal{X}} |\mathcal{S}_{\xseq}|.
  \label{eq:agop}
\end{equation}
This averages the gradient outer products equally across all \(N\) selected positions in \(\mathcal{X}\) (see \cref{subsec:probe-setting} for the evaluation corpus and position-sampling details).

\subsection{The alignment between concept subspace and loss geometry}
\label{subsec:optim-engagement-def}

Given a concept subspace with orthonormal basis \(\Umat_{\ell,t}\in\mathbb{R}^{d\times k_{\ell,t}}\), we measure what fraction of the total squared norm of the loss gradients is captured by this subspace,
\begin{equation}
    \pi_{\ell,t} = \frac{\sum_{\xseq\in\mathcal{X}}\sum_{s\in\mathcal{S}_{\xseq}}\|\Umat_{\ell,t}^\top \gvec_{\ell,t}(\xseq,s)\|_2^2}
       {
        \sum_{\xseq\in\mathcal{X}}\sum_{s\in\mathcal{S}_{\xseq}}\|\gvec_{\ell,t}(\xseq,s)\|_2^2
       }
       = \frac{\tr(\Umat_{\ell,t}^{\top}\Gmat_{\ell,t}\Umat_{\ell,t})}{\tr(\Gmat_{\ell,t})},
   \end{equation}
where both sums run over the $N$ selected positions in \cref{eq:agop}, and \(\tr(\cdot)\) denotes the trace of a matrix.

A uniformly random subspace of the same rank as the concept subspace captures, in expectation, a \(\rk(\Umat_{\ell,t})/d\) fraction of the trace of any fixed positive-semidefinite matrix with positive trace (see \Cref{app:haar-expectations}).
We therefore define \oetext as the log-fold enrichment of the observed trace share over this matched-rank random baseline:
\begin{equation}
  \oem_{\ell,t}
  =
  \log_2
  \frac{
  \pi_{\ell,t}
  }{
  \rk(\Umat_{\ell,t})/d
  }.
  \label{eq:optim-engagement}
\end{equation}
Thus \(\oem_{\ell,t}=0\) indicates that the observed trace share equals the expected trace share of a uniformly random subspace of the same rank; positive values indicate a larger share, and one unit corresponds to a factor-of-two increase in enrichment. 

\subsection{\cetext: concept-specific optimization engagement}
\label{subsec:concept-energy-def}

A well-known problem with probes is that they may achieve high accuracy during training by memorizing associations between token types and outputs. 
Accordingly, \citet{hewitt-liang-2019-designing} propose control tasks, which associate word types with random outputs to measure a probe's selectivity for the concept of interest.
Similarly, \(\oem\) itself is not concept-specific, as it may be affected by the model's changing ability to memorize such associations.
Therefore, we define \cetext as the difference in \(\oem\) for a real task and a control task,
\begin{equation}
    \ce = \oem_{\text{real}} - \oem_{\text{ctrl}}.
    \label{eq:concept-energy-def}
\end{equation}
Only \(\ce > 0\) indicates concept-specific alignment between the concept subspace and the loss geometry.
We discuss the differences between using \(\oem\) and \(\ce\) to track concept learning trajectories in detail in \Cref{app:subsec:oe-control-ce}.

\section{Experimental methods}
\label{sec:experimental-methods}

\subsection{Probe setting}
\label{subsec:probe-setting}

The analysis is conducted on Pythia-1B and -410M at 19 approximately log-linearly spaced training checkpoints from step 1 to step \(143{,}000\).
For each layer at each checkpoint, we run both real and control tasks with three random seeds.
The detailed configurations are in \Cref{app:probe-details}.
The probe suite consists of 12 tasks grouped into three families:
\begin{itemize}[leftmargin=2em,topsep=2pt,itemsep=2pt]
  \item \emph{Lexical statistics.} \unigramlp (unigram log-probability), \bigramlp and \trigramlp (conditional bigram and trigram log-probabilities), and \bigrampmi and \trigrampmi (bigram and trigram pointwise mutual information).
  The n-gram statistics that serve as regression targets are computed from WikiText-103~\citep{merity2017pointer} under the model's BPE tokenization, and probe fits are evaluated by held-out \(R^2\).\footnote{%
        Throughout the paper, ``unigram'', ``bigram'', and ``trigram'' refer to token n-grams under the Pythia BPE tokenization, not whitespace-delimited words.
        In addition, we center all n-gram target statistics when training the n-gram probes; see \Cref{app:centring} for details.
    }
    The motivation for including these regression targets and their derivation is provided in \Cref{app:motive-analysis-ngram} and \Cref{app:chain-rule-proof}, respectively.

  \item \emph{Syntax.} \pos (universal POS tag), \deplen (local dependency arc length), \deprel (dependency relation label), and \depdepth (discretized depth-to-root). Labels come from Universal Dependencies (UD) English EWT~\citep{silveira-etal-2014-gold,nivre-etal-2016-universal}, and probes are evaluated by accuracy. For the four syntactic tasks, punctuation tokens are excluded from probe training and evaluation, though they are still provided in the input to the model.
  \item \emph{Semantics.} \ner (coarse named-entity tag), \ssnoun (single-word noun supersense), and \ssverb (single-word verb supersense). NER labels come from the Universal NER (UNER) English EWT release~\citep{mayhew-etal-2024-universal}; the supersense labels come from STREUSLE~\citep{schneider-smith-2015-corpus,schneider-etal-2018-comprehensive}, which extends UD EWT with single-word noun and verb supersense annotations. All three are evaluated by accuracy.
\end{itemize}

Control tasks follow \citet{hewitt-liang-2019-designing}: each word type is assigned a label drawn from the token-level distribution of real labels.
For the n-gram tasks, a type is defined as a single word type, the \texttt{(previous, current)} pair, or the \texttt{(previous two, current)} triple.
Control tasks have the same label distribution as the corresponding target task.
Probe activations are extracted from sequences truncated to 256 tokens; for AGOP estimation, sequences are truncated to 128 tokens.
To estimate AGOP, we sample 1{,}000 evaluation sequences from WikiText-103. From each sequence, we uniformly sample up to eight eligible next-token positions without replacement, and average the gradient outer products equally across all sampled positions.
When calculating $\oem_{\text{real}}$ and $\oem_{\text{ctrl}}$ in \cref{eq:concept-energy-def}, we average each over the three seeds.
Probe training protocols, dataset statistics and pre-processing are detailed in \Cref{app:probe-details}.

\subsection{Event timing}
\label{subsec:event-timing}

To quantify the temporal relation between \cetext and changes in decodability, we compare two events:
(a) the fastest observed improvement in the evaluation measure
of interest, and (b) the earliest qualifying peak in \cetext.
We use the earliest qualifying peak because the precedence hypothesis concerns the first episode of concept-specific engagement; later peaks are discussed in \Cref{subsec:multipeak}. 

For a higher-is-better evaluation curve $M$ measured at checkpoints $t_1 < \cdots < t_N$, we operationalize (a) using
a common transition-event definition:
\begin{equation}
  i^\star(M)
  = \operatorname*{arg\,max}_{1 \le i < N}
     \frac{M(t_{i+1})-M(t_i)}
          {\log t_{i+1}-\log t_i},
    \qquad
  t^{M}
  = \sqrt{t_{i^\star(M)}\,t_{i^\star(M)+1}}.
  \label{eq:transition-event}
\end{equation}

Thus, $t^{M}$ is the midpoint in log time of the interval with the largest observed increase per unit log training step.
Ties are resolved by selecting the earliest interval;
the event is undefined if the maximum slope is nonpositive.
For decodability, we apply this definition to the selectivity curve at each layer, denoting it as $t^D_\ell$.
We operationalize (b) as the earliest non-spurious local maximum of the \cetext trajectory, denoted by $t^E_\ell$.\footnote{%
    As small fluctuations in the curve can lead to spurious local maxima, we apply a more stringent criterion to select $t^{E}_{\ell}$. 
    See \Cref{app:peak-rule} for details.
}

To obtain a global event time for the entire model for a given task, we aggregate over all layers. 
Let $\mathcal V$ denote the layers with both a defined energy event and a defined decodability transition.
We aggregate decodability and energy event times as
\begin{equation}
  t^D = \exp\left(
    \frac{\sum_{\ell\in\mathcal V}w_\ell\log t_\ell^D}
         {\sum_{\ell\in\mathcal V}w_\ell}
  \right),
  \,
  t^E = \exp\left(
    \frac{\sum_{\ell\in\mathcal V}w_\ell\log t_\ell^E}
         {\sum_{\ell\in\mathcal V}w_\ell}
  \right),
  \,
  w_\ell=\max_i[\ce_{\ell,t_i}]_+.
  \label{eq:aggregated-event-time}
\end{equation}
In practice, we compute \(t^D\) from the selectivity curve, defined as the difference between the held-out scores averaged across three real-task seeds and three control-task seeds.
We compare \(t^D\) and \(t^E\) to determine the temporal ordering between the decodability and \cetext events.
In particular, when $\frac{\sqrt{2}}{2} < \frac{t^E}{t^D} < \sqrt{2}$,
i.e., when the two events are separated by less than half a checkpoint interval, we treat their relative order as indistinguishable and refer to them hereafter as coinciding.\footnote{%
The number of steps approximately doubles between successive checkpoints.
}

\section{Results}
\label{sec:results}

\subsection{Concept Energy is positive for most tasks and layers}
\label{subsec:ce-positive}

\begin{figure}[!t]
  \centering
  \includegraphics[width=\linewidth]{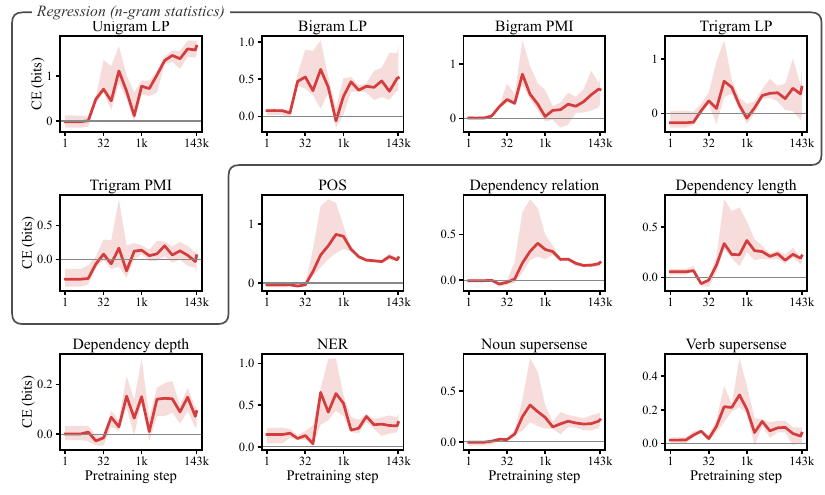}
  \caption{
    \cetext over pretraining for the 12 tasks in Pythia-1B.
    Lines and bands: median and 25th--75th percentiles across the 16 layers.
    The outlined panels are the five regression (n-gram) tasks; the other seven are classification tasks.
    Y-axes differ across panels.
  }
  \label{fig:ce-timecourse-1b}
\end{figure}

\Cref{fig:ce-timecourse-1b} shows how \cetext changes over pretraining across all 12 tasks for Pythia-1B.%
\footnote{Corresponding results for Pythia-410M are in \Cref{fig:ce-timecourse-410m}.}
For \emph{classification} tasks, \cetext mostly peaks between steps 64 and 1k; \pos, \deprel, \deplen, \ner, \ssnoun, and \ssverb decline after peaking but remain positive, whereas \depdepth continues to fluctuate in later stages.
\emph{Regression} tasks exhibit greater fluctuation: \trigrampmi remains close to 0 throughout;\footnote{%
    Although for \trigrampmi, the five shallowest layers of Pythia-1B each have a single peak at step 128 (\Cref{fig:ce-heatmap-1b}).
}
the other n-gram tasks show varying degrees of upward trends in later training.
\Cref{tab:ce-positive-share} reports the share of (layer, checkpoint) measurements with \cetext$>0$ across steps 1--32 and steps 64--143k, respectively.
During the first 32 steps, \cetext stays near 0, with many measurements near or below 0 and only 59\% of measurements positive across all tasks (55\% in Pythia-410M); \unigramlp and \bigramlp are exceptions, which rise from step 16 onward.
From step 64 onward, the majority of measurements are positive (90\%; 87\% in Pythia-410M).
In both models, 11 of 12 tasks have a higher positive share in the later phase; the exceptions are \ner in Pythia-1B (100\% vs.\ 98\%) and \depdepth in Pythia-410M (78\% vs.\ 76\%).

\begin{table}[!t]
  \centering\small
  \setlength{\tabcolsep}{5pt}
  \caption{
    Share (\%) of (layer, checkpoint) cells with \(\ce>0\) at steps 1--32 and at steps 64--143k (\(\geq\)64), for the regression tasks (left) and the classification tasks (right); the last row pools the cells of all 12 tasks.
    The two groups are split where the pooled share rises most: from 67\% at step 32 to 85\% at step 64 in Pythia-1B, and from 64\% to 78\% in Pythia-410M.
  }
  \label{tab:ce-positive-share}
  \begin{tabular}{lrrrr@{\hspace{1.5em}}lrrrr}
\toprule
\multirow{2}{*}[-2.5pt]{\emph{Regression}} & \multicolumn{2}{c}{Pythia-1B} & \multicolumn{2}{c}{Pythia-410M} & \multirow{2}{*}[-2.5pt]{\emph{Classification}} & \multicolumn{2}{c}{Pythia-1B} & \multicolumn{2}{c}{Pythia-410M} \\
\cmidrule(lr){2-3}\cmidrule(lr){4-5}\cmidrule(lr){7-8}\cmidrule(lr){9-10}
 & 1--32 & $\geq$64 & 1--32 & $\geq$64 & & 1--32 & $\geq$64 & 1--32 & $\geq$64 \\
\midrule
\unigramlp & 64 & 98 & 56 & 95 & \pos & 24 & 100 & 80 & 100 \\
\bigramlp & 82 & 92 & 74 & 90 & \deprel & 32 & 97 & 49 & 91 \\
\bigrampmi & 75 & 85 & 71 & 72 & \deplen & 65 & 98 & 74 & 93 \\
\trigramlp & 50 & 80 & 31 & 76 & \depdepth & 45 & 87 & 78 & 76 \\
\trigrampmi & 22 & 62 & 33 & 69 & \ner & 100 & 98 & 26 & 96 \\
 & & & & & \ssnoun & 62 & 100 & 20 & 98 \\
 & & & & & \ssverb & 82 & 90 & 69 & 84 \\
\midrule
All 12 tasks & 59 & 90 & 55 & 87 & & & & & \\
\bottomrule
\end{tabular}

\end{table}

\subsection{Energy usually rises before decodability, but not universally}
\label{subsec:timing-diagonal}

\begin{figure}[!tbp]
  \centering
  \includegraphics[width=\linewidth]{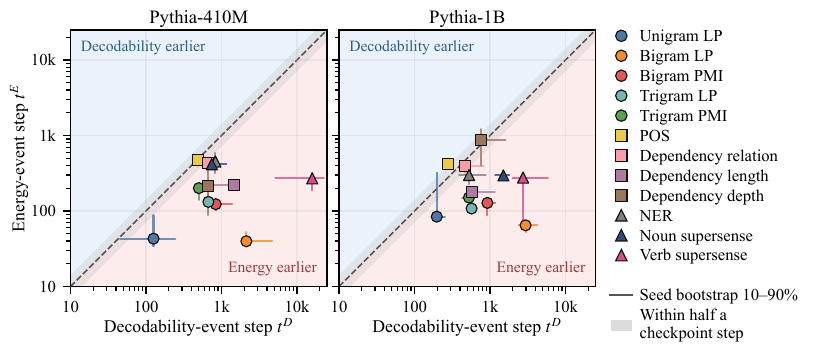}
  \caption{
    Each point aggregates layer-wise decodability and energy events using weighted geometric means.
    Layer weights are defined in \cref{eq:aggregated-event-time}.
    Points below the diagonal indicate earlier energy events.
    When a task falls within the gray band around the diagonal, we consider two events to coincide for that task, as the band represents a separation of less than half a checkpoint interval ($\frac{\sqrt{2}}{2} < \frac{t^E}{t^D} < \sqrt{2}$).
  }
  \label{fig:probe-decode-diagonal}
\end{figure}

The geometric-precedence hypothesis predicts that the \cetext peak will generally precede the steepest transition in decodability.
Our results broadly support this hypothesis.
In \Cref{fig:probe-decode-diagonal}, each point summarizes valid layers using
\Cref{eq:aggregated-event-time}; 
points falling below the diagonal indicate that the energy event occurred earlier.
Energy events precede decodability transitions in 11 of 12 tasks in Pythia-410M, while the two events coincide for \pos; in Pythia-1B, energy events likewise precede decodability transitions in 9 of 12 tasks, coincide for \deprel and \depdepth, and occur slightly after the decodability transition for \pos.
We refer to the numbers of tasks in which the energy event occurs earlier than, coincides with, or occurs later than the decodability event as \emph{timing-order counts}, written hereafter as \texttt{earlier/same/later}; for example, in \Cref{fig:probe-decode-diagonal}, the timing-order counts are \texttt{11/1/0} for Pythia-410M and \texttt{9/2/1} for Pythia-1B.
As the aggregation may hide opposing temporal lag effects between layers, we provide the full trajectories with event times in \cref{app:subsec:trajectories} and detailed layer-wise timing statistics in \cref{app:subsec:temporal-summary-order-sensitivity}.

\subsection{Probe-independent behavior}
\label{subsec:behavior}

The impact of a concept on model behavior is not always predictable from the probe score \citep{vig-etal-2020-causal-mediation-analysis}.
To examine the relationship between \cetext and model behavior, we train a subject--verb agreement (SVA) number probe on Pythia-410M and compare probe performance and \cetext to a behavioral evaluation of SVA.
Given a prefix carrying grammatical number information, this probe predicts the number of the immediately following verb.
Specifically, we train a binary classification probe on the SVA dataset of \citet{linzen-etal-2016-assessing}, using the subset released with \citet{goldberg2019assessing}, where each sentence contains 1--4 nouns with the opposite number from the subject (attractors) intervening between the subject and the verb.
At each layer, the probe reads the final token representation of the pre-verbal prefix and predicts whether the upcoming verb is singular or plural.
Control task labels are tied to the final token of the prefix.\footnote{%
    A lexical baseline that inspects only the final token of the prefix already achieves a held-out accuracy of 0.847 (majority-class baseline is 0.669); the real probe rises to 0.93 after roughly 1k steps, surpassing this baseline.
}
We use the model's macro-average accuracy across the six paradigms in the SVA category of BLiMP~\citep{warstadt-etal-2020-blimp-benchmark} as the \emph{behavioral} measure.
We treat this behavioral measure as the evaluation curve in \cref{eq:transition-event} to obtain the behavioral transition event $t^B$.
We compute the decodability transition event and the earliest qualifying \cetext peak for SVA and compare them against the BLiMP-derived $t^B$.

\begin{figure}[!tbp]
  \centering
  \includegraphics[width=\linewidth]{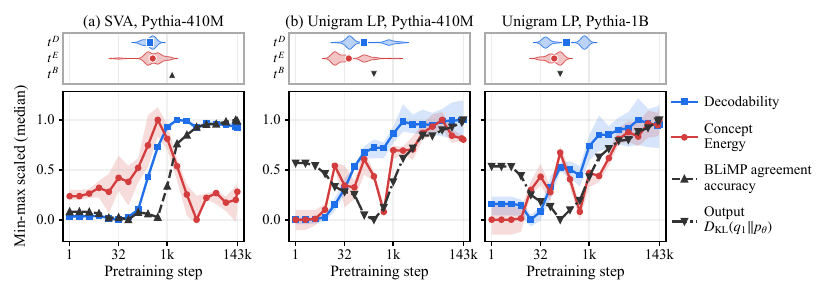}
  \caption{
    Decodability, \cetext, and probe-independent behavior over pretraining.
    (a) Subject--verb agreement in Pythia-410M; behavior is BLiMP agreement accuracy.
    (b) \unigramlp in Pythia-410M (left) and Pythia-1B (right); behavior is the divergence $D_{\mathrm{KL}}(q_1\|p_\theta)$ between the WikiText unigram distribution and the model's next-token distribution (lower is closer).
    Lines are medians over layers, min--max scaled; bands span the nine pairs of real- and control-probe seeds.
    The strip above each panel marks $t^D$, $t^E$, and $t^B$ on the same step axis.
    $t^D$ and $t^E$ are aggregated over layers by \cref{eq:aggregated-event-time}; their violins show the $w_\ell$-weighted distribution of the layer events, so the aggregate need not coincide with features of the plotted medians: in Pythia-1B, for example, \unigramlp decodability rises most steeply between steps 32 and 64 in five of the 14 valid layers and between steps 512 and 1{,}000 in eight, so $t^D$ falls between the two.
    $t^B$ is the steepest rise in accuracy in (a) and the minimum divergence in (b); it has no violin because (a) is a single evaluation and in (b) the minimum falls on the same checkpoint in all 1{,}000 bootstrap resamples of the documents.
  }
  \label{fig:sva-decode-energy-behavior}
\end{figure}

As shown in \Cref{fig:sva-decode-energy-behavior}, $t^D=297$, $t^E=357$, and $t^B=1414$: the decodability transition event and the energy event coincide ($t^E/t^D\approx 1.20<\sqrt{2}$).
Both the energy and decodability events precede the behavioral transition by about two checkpoint intervals ($1414/357\approx3.96$).
This provides an example in which loss geometry precedes model behavior.

Additionally, we compare the model's output distribution with a fixed n-gram reference distribution following \citet{chang-bergen-2022-word}.
For Pythia-410M/-1B models, we apply this diagnostic to unigram, bigram, and trigram reference distributions.
At each selected checkpoint, we calculate \(D_{\mathrm{KL}}(q_n\|p_\theta)\), where \(q_n\) is the reference n-gram distribution constructed from WikiText and \(p_\theta\) is the model's next-token prediction distribution.
Lower divergence indicates greater similarity between the model's output distribution and the reference.
Unlike accuracy, $D_{\mathrm{KL}}(q_n\|p_\theta)$ is neither higher- nor lower-is-better: models first approach $q_n$ and then move away~\citep{chang-bergen-2022-word}. 
Following \citet{chang-bergen-2022-word,chang-etal-2024-characterizing}, we define the behavioral event $t^B$ as the checkpoint of minimum divergence, where the model's predictions are closest to the n-gram reference.

Changes in the decodability of n-gram log conditional probabilities do not necessarily capture changes in the model's output.
The divergence between the model's output distribution and the reference distribution exhibits a trend of initially decreasing and then increasing; conversely, probe performance remains essentially stable after reaching a certain level.
Taking the unigram case as an example, as shown in \Cref{fig:sva-decode-energy-behavior}(b),\footnote{%
    Bigram and trigram results for both models are in \Cref{app:behavior}.
} the minimum divergence between the outputs of the Pythia-410M and Pythia-1B models and the reference on WikiText occurs at step 256 and step 128, respectively.
Although the divergence increases subsequently, the associated statistical features retain a high level of decodability.
Across both models and all three orders, the energy event precedes the divergence minimum in all six cases (\Cref{app:behavior}).

\subsection{Robustness}
\label{subsec:robustness}

To assess the robustness of the analytical framework presented in this work, we vary one analysis choice at a time (details in \Cref{app:probe-reliability,app:centring}).
Specifically,
\begin{enumerate}[
    leftmargin=1.5em,
    labelsep=0.3em,
    itemsep=0pt
]
    \item \textbf{Independently fitted probes.} 
    In \cref{sec:theoretical-method}, we use the same set of real and control probes to compute both \cetext and selectivity; to test whether concept-specific engagement arises merely from this reuse, we re-fit the POS probes of Pythia-410M on two disjoint halves of the sentences (A and B).
    Then we identify the energy event from one half and the decodability event from the other.
    We test both using A for decodability and B for the energy event and vice versa; in both directions, the decodability transition event and the energy event differ by less than half a checkpoint interval, i.e., the two events coincide.
    Details are in \Cref{app:independent-probes}.
    \item \textbf{Number of probe directions.}
    To assess the sensitivity of concept-specific engagement to the number of directions used to construct the subspace, we re-compute POS \cetext for Pythia-410M for $k = 1, 4, 8, 12$ (full probe: $k=15$).
    We find that for $k\ge 4$, the median Spearman $\rho$ between each layer's \cetext trajectory and the full $k=15$ trajectory is at least 0.93.
    This indicates that the shape of the \cetext curve is largely insensitive to the number of directions.
    Details are in \Cref{app:k-sensitivity}.
    \item \textbf{Probe-training budget.} 
    Some probing tasks have limited labeled data.
    To evaluate the sensitivity of \oetext and \cetext when probes are trained on less data and fit less well, we train POS probes for Pythia-410M on $1/4, 1/16 \text{, and } 1/64$ of the data at layers 5, 13, and 21. 
    We find that for probes trained on reduced data, peak \(\ce\) is $0.98$--$1.14\times$ its full-data value (median across the three layers), while peak \(\oem\) is $1.7$--$3.1\times$ its full-data value.
    This indicates that \oetext itself is sensitive to fit quality, while \cetext mitigates this effect by subtracting the control task's \(\oem\).
    Details are in \Cref{app:probe-data-sensitivity}.
    \item \textbf{Input centering.} 
    Because shared mean directions in the activations may affect the probe weight matrix, we evaluate whether the shape of the \cetext curve is influenced by these shared directions.
    Without mean-centering the probe inputs (Pythia-410M, 12 tasks), the per-task Spearman $\rho$ with the \cetext of probes trained on mean-centered inputs is $0.81$--$0.99$, and the timing-order counts are unchanged.
    This indicates that the observed concept-specific engagement pattern is not driven by shared mean directions.
    Details are in \Cref{app:centring}.
\end{enumerate}

\section{Discussion}
\label{sec:discussion}

\subsection{What \cetext adds beyond decodability and behavior}
\label{subsec:discussion-what-oe-add-beyond}

Probe performance tells us when a concept is decodable from a model, while a model's behavior tells us when its acquisition of a concept manifests in its output.
However, an increase in either of these measures alone does not establish that the training objective is currently pressuring the development of the concept.
\cetext asks the complementary question of whether the next-token loss geometry is captured by the concept subspace, and it can be high while decodability is low, rising, converged, or a mixture of these states across layers.

\subsection{\cetext can have more than one peak}
\label{subsec:multipeak}

Although \Cref{subsec:timing-diagonal} shows that for Pythia-410M and -1B, most tasks satisfy the conclusion that geometry precedes decodability, and \Cref{subsec:behavior} shows that geometry also precedes behavior for the SVA and n-gram tasks, both analyses use the earliest qualifying peak as the energy event.
However, the \cetext curve does not always exhibit a single, distinct peak.
In some cases, \cetext remains high after reaching the earliest qualifying peak, or (as in the \unigramlp example in \Cref{fig:example-decodability-vs-energy}(a)) it drops and then rises again, resulting in multiple peaks.
These fluctuations may persist even after the decodability curves have stabilized.
Specifically, across all tasks, 44.8\% of layer trajectories in Pythia-410M have more than one peak, and 45.8\% in Pythia-1B.
We report complete per-task multi-peak descriptive statistics for both models in \Cref{app:subsec:multipeak}.

In \Cref{subsec:event-timing,subsec:timing-diagonal}, we use the earliest qualifying peak as the \cetext event, which yields timing-order counts of \texttt{11/1/0} (410M) and \texttt{9/2/1} (1B). 
In fact, alternative definitions of the energy event lead to different timing results.
When the energy event is defined as the global maximum peak ($\tau_{\max}$), the timing-order counts are \texttt{8/2/2} (410M) and \texttt{7/3/2} (1B); when defined as the step at which the cumulative positive \cetext area reaches half ($\tau_{50}$), they are \texttt{4/5/3} (410M) and \texttt{4/5/3} (1B). 
We provide a full comparison of all three energy-event definitions in \Cref{app:probing-timing-analyses}.
Additionally, we present the complete evolutionary trajectories of these metrics across different tasks in \Cref{fig:ce-heatmap-410m,fig:ce-heatmap-1b}.

\subsection{Limitations}
\label{subsec:limitations}

First, \cetext measures loss-gradient geometry, not causal reliance; it should not be cited as evidence that the model uses a concept.
Second, probe subspaces are operational definitions: they can contain correlates of the target label and can miss nonlinear concept structure.
Third, the main analysis is limited to two sizes within the Pythia family and should be replicated across scale, architecture, and data order.
Fourth, because not all tasks have a corresponding behavioral benchmark, we do not test the temporal ordering between energy events and behavioral transitions across all tasks.

\section{Conclusion}
\label{sec:conclusion}

We introduced \cetext for tracking concept learning during language-model pretraining.
In Pythia-410M and -1B, positive \cetext can be observed across 12 lexical, syntactic, and semantic probing tasks, indicating concept-specific engagement with these concepts during pretraining.
Moreover, for most tasks across both models, the earliest \cetext events precede decodability transitions, broadly supporting the hypothesis that geometry precedes decodability.
Furthermore, \citet{chang-bergen-2022-word}-style analyses on n-grams show that high decodability can persist even as model predictions diverge from fixed n-gram reference distributions, and energy events precede the point at which model predictions are closest to the n-gram reference for both models and all three n-gram orders.
Complementary SVA results also suggest that geometry precedes model behavior.
Taken together, these findings distinguish concept-specific optimization engagement from decodability and model behavior, providing a complementary view of concept development.
Future work should examine the robustness of these temporal patterns across model scales, architectures, and data orders.

\begin{ack}

The authors thank Kyle Mahowald, Sasha Boguraev, and Kaijie Mo for discussions and inspiration. During the experiments, the authors used the AI coding agents Codex and Claude Code to check mathematical derivations, find bugs in code, and orchestrate large-scale probe training. During writing, Claude Opus 5.5 was used to read drafts, suggest revisions, and produce some of the figures.

This work used Jetstream2 at Indiana University through allocation CIS220009 from the Advanced Cyberinfrastructure Coordination Ecosystem: Services \& Support (ACCESS) program, which is supported by U.S. National Science Foundation grants \#2138259, \#2138286, \#2138307, \#2137603, and \#2138296~\citep{boerner2023access}.
Early exploratory experiments were run on the National Research Platform. This work used resources available through the National Research Platform (NRP) at the University of California, San Diego. NRP has been developed, and is supported in part, by funding from National Science Foundation, from awards 1730158, 1540112, 1541349, 1826967, 2112167, 2100237, and 2120019, as well as additional funding from community partners~\citep{weitzel2025nrp}.

\end{ack}

\bibliographystyle{plainnat}
\bibliography{references}

\appendix
\crefalias{section}{appendix}
\crefalias{subsection}{appendix}

\section{Probe training details}
\label{app:probe-details}

We use classification probes for \pos, \deplen,
\deprel, \depdepth, \ner, \ssnoun,
and \ssverb, and regression probes for the five
lexical-distributional tasks.
All probes are linear, trained with Adam at learning rate \(10^{-3}\) and weight decay \(10^{-4}\) on layer-wise residual-stream activations from frozen Pythia-1B and -410M checkpoints.

\paragraph{Datasets.} The syntactic tasks use UD English EWT v2.17~\citep{silveira-etal-2014-gold,nivre-etal-2016-universal}.
The supersense tasks use STREUSLE 5.0~\citep{schneider-smith-2015-corpus,schneider-etal-2018-comprehensive}, whose syntactic layer is the same EWT v2.17 release.
\ner uses UNER~\citep{mayhew-etal-2024-universal}, which was annotated on the earlier EWT v2.9.\footnote{We use commit \texttt{8ed072de2fd24022cc62458997fe96a8fe191ea4} of \url{https://github.com/UniversalNER/UNER_English-EWT}; its annotations are aligned to UD English EWT commit \texttt{c491d62a7a0447946ac85b11308083ea61328ece}, identical to release v2.9.}

\paragraph{Classification probes.}
We use 10 epochs, batch size 256, and softmax cross-entropy.
We use all sentences from the official training and test splits that pass the sentence and label filters, without imposing a fixed sentence or token cap.
We retain sentences with at least four tokens and two alphabetic tokens, exclude sentences containing URL-like tokens, and truncate model inputs to 256 BPE tokens.
For every classification task we keep the classes with at least 100 training and 10 test tokens.
The seven classification tasks differ in their target label sets and class counts.
\begin{itemize}[leftmargin=2em,topsep=2pt,itemsep=2pt]
  \item \pos: 16 universal POS tags from UD (\texttt{ADJ}, \texttt{ADP}, \texttt{ADV}, \texttt{AUX}, \texttt{CCONJ}, \texttt{DET}, \texttt{INTJ}, \texttt{NOUN}, \texttt{NUM}, \texttt{PART}, \texttt{PRON}, \texttt{PROPN}, \texttt{SCONJ}, \texttt{SYM}, \texttt{VERB}, \texttt{X}; \texttt{PUNCT} is excluded).
  \item \deplen: 9 signed-distance bins encoding head--dependent direction and arc length: \texttt{ROOT}, \texttt{LEFT\_1}, \texttt{LEFT\_2}, \texttt{LEFT\_3\_5}, \texttt{LEFT\_6P}, and the symmetric \texttt{RIGHT} bins.
  \item \deprel: 30 universal dependency relation labels from UD, retained from the 35 relations present in EWT after applying the $\ge 100$ train and $\ge 10$ test token filter.
  \item \depdepth: 12 discretized depth-to-root classes (\texttt{0}--\texttt{10} and \texttt{11+}), obtained from the 17 raw depth bins (\texttt{0}--\texttt{16}) by merging all depths $\ge 11$ into \texttt{11+}.
  \item \ner: 3 coarse named-entity classes from UNER (\texttt{PER}, \texttt{ORG}, and \texttt{LOC}), obtained from 4 original tags by excluding the non-entity tag \texttt{O}.
  \item \ssnoun: 12 WordNet-derived noun supersenses from STREUSLE (\texttt{n.PERSON}, \texttt{n.LOCATION}, \texttt{n.ARTIFACT}, \texttt{n.COMMUNICATION}, \(\ldots\)), retained from the 26 original classes under the minimum-count filter.
  \item \ssverb: 9 WordNet-derived verb supersenses from STREUSLE (\texttt{v.communication}, \texttt{v.motion}, \texttt{v.cognition}, \texttt{v.change}, \(\ldots\)), retained from the 14 original classes under the minimum-count filter.
\end{itemize}

\paragraph{Regression probes.}
We train the linear regression probes for 10 epochs with a batch size of 256, and squared-error loss.
Sentences from WikiText-103~\citep{merity2017pointer} are tokenized with the Pythia BPE tokenizer and truncated to 256 tokens, keeping only lines with at least four tokens and two alphabetic tokens while excluding lines containing URL-like tokens.
From 20{,}000 filtered sentences sampled with seed 42, we read sentences in order up to 5{,}000 sentences or 50{,}000 token positions for the unigram and bigram tasks, and up to all 20{,}000 sentences or 500{,}000 token positions for the trigram tasks.

Each task is split by its own n-gram types: single tokens for unigrams, $(x_{t-1},x_t)$ pairs for bigrams, and $(x_{t-2},x_{t-1},x_t)$ triples for trigrams.
Among types with count $\ge 2$, we randomly assign 80\% to training and 20\% to the test set with seed 42.

For bigram and trigram tasks, we further filter the test set to retain only n-grams whose lower-order components appear in the training split (both $x_{t-1}$ and $x_t$ for bigrams; both the prefix $(x_{t-2},x_{t-1})$ and $x_t$ for trigrams), leaving 451 of 1{,}035 held-out bigram types and 3{,}004 of 7{,}518 held-out trigram types.
The resulting train / held-out token counts are 37{,}287 / 7{,}996 for \unigramlp, 16{,}685 / 2{,}815 for the bigram tasks, and 108{,}566 / 12{,}154 for the trigram tasks.

We use no smoothing for bigram or trigram probabilities and fit probes only on n-grams with non-zero counts in the WikiText-103 reference statistics;
control targets are drawn per n-gram type within the training and test sets separately, and targets are centered by the training-set mean (\Cref{app:centring}).

\paragraph{Control tasks.}
We follow \citet{hewitt-liang-2019-designing} to design control tasks for both classification and regression tasks.
For \emph{classification}, we randomly assign each word type a label drawn from the token-level label distribution of the real task.
Since the training and test splits share word types, the control probe can achieve a high score on the test set by memorizing word identities.
For the \emph{n-gram regression} tasks, we also randomly assign each n-gram type a target statistic drawn from the token-level target distribution (within the training and test sets separately).
Therefore, because the training and test sets have zero n-gram type overlap, merely memorizing n-gram types in the training set will not enable the control probe to generalize.
By contrast, the real-task probe can still generalize because the statistic is encoded in representations shared across n-gram types; for the bigram and trigram tasks, every test-set n-gram is composed of tokens that appeared in the training set.

\paragraph{Decodability metric.}
The held-out accuracy and \(R^2\) values used to compute decodability in the main text are evaluated on each task's held-out split.
Selectivity is then computed by first averaging the held-out probe scores across seeds for the real task and the control task separately, and taking their difference.

\paragraph{Evaluation checkpoints.}
The 19 evaluation checkpoints described in \Cref{subsec:probe-setting} are \(\mathcal{S} = \{1,\ 2,\ 4,\ 8,\ 16,\ 32,\ 64,\ 128,\ 256,\
512,\ 1\mathrm{k},\ 2\mathrm{k},\ 4\mathrm{k},\ 8\mathrm{k},\ 16\mathrm{k},\ 32\mathrm{k},\ 64\mathrm{k},\ 128\mathrm{k},\ 143\mathrm{k}\}\); the first ten are powers of two and the last nine are approximately log-spaced round-number steps to step \(143\mathrm{k}\).
The final interval, from step \(128\mathrm{k}\) to \(143\mathrm{k}\), is shorter on the log-time axis.

\section{N-gram targets and mathematical derivations}
\label{app:math-derivations}

\subsection{Motivation and analysis of the n-gram regression tasks}
\label{app:motive-analysis-ngram}

The choice of n-gram statistics as probe targets is motivated by prior findings that LMs acquire these statistics in a structured temporal order during pretraining, with unigram and short-context predictions learned before longer-context ones~\citep{chang-etal-2024-characterizing}.
Our decomposition lets us ask which component of this acquisition the model's internal geometry tracks at each stage.

These five lexical-statistic targets are the components of a chain-rule decomposition of the next-token log-probability
(see the derivation in \Cref{app:chain-rule-proof}).
For our bigram and trigram conditional probes, this gives the explicit decompositions
\begin{align}
  \log p(x_t \mid x_{t-1}) &= \log p(x_t) + \PMI(x_t;\, x_{t-1}),
  \notag\\
  \log p(x_t \mid x_{t-2},\,x_{t-1}) &= \log p(x_t \mid x_{t-1}) + \PMI(x_t;\, x_{t-2}\mid x_{t-1}).
  \label{eq:ngram-decomposition}
\end{align}
Thus, while a probe can partly fit \bigramlp and \trigramlp (\(\log p(x_t \mid x_{t-1})\) and \(\log p(x_t \mid x_{t-2},\,x_{t-1})\), respectively) by capturing the lower-order terms such as \(\log p(x_t)\), fitting \bigrampmi and \trigrampmi, i.e., \(\PMI(x_t;\,x_{t-1})\) and \(\PMI(x_t;\,x_{t-2}\mid x_{t-1})\), requires recovering the higher-order PMI components themselves.

\Cref{eq:ngram-decomposition} therefore suggests that pure-PMI targets should be harder to decode than mixed targets that include the lower-order base, with the unigram base itself the easiest.
The best-across-layers probe \(R^{2}\) at the final checkpoint (step \(143{,}000\)) is shown in \Cref{tab:ngram_prob_performance_comparison}.
\begin{table}[t]\centering\small
\caption{Test-set $R^2$ of linear probes on the five n-gram targets at the final checkpoint (step 143,000).
Values show the best seed-mean $R^2$ across layers, with the best layer (1-indexed) in parentheses.}
\begin{tabular}{lrr}
\toprule
Task & Pythia-1B & Pythia-410M \\
\midrule
Unigram LP & 0.886 (5) & 0.890 (5) \\
Bigram LP & 0.824 (4) & 0.835 (3) \\
Bigram PMI & 0.625 (4) & 0.644 (3) \\
Trigram LP & 0.764 (4) & 0.758 (6) \\
Trigram PMI & 0.179 (5) & 0.157 (6) \\
\bottomrule
\end{tabular}

\label{tab:ngram_prob_performance_comparison}
\end{table}
In both models, unigram log-probability is the most decodable target, while the PMI targets have much lower best-layer $R^2$ than their corresponding conditional log-probability targets, especially for trigrams.
This contrast motivates analyzing the higher-order components separately rather than treating conditional-probability decodability as evidence that each component is equally recoverable.

\subsection{Chain-rule decomposition of next-token log-probability}
\label{app:chain-rule-proof}

\Cref{eq:ngram-decomposition} writes the next-token log-probability as a unigram base term plus a series of higher-order PMI terms.
We give the elementary derivation here.
For any joint distribution over \(x_1,\ldots,x_t\) and any \(1\leq k\leq t-1\), the conditional log-probability satisfies
\begin{align*}
  & \log p(x_t\mid x_{t-k},\ldots,x_{t-1})
  - \log p(x_t\mid x_{t-k+1},\ldots,x_{t-1}) \\
  =\;& \log\frac{p(x_t\mid x_{t-k},\ldots,x_{t-1})}{p(x_t\mid x_{t-k+1},\ldots,x_{t-1})} \\
  =\;& \log\frac{p(x_t,x_{t-k}\mid x_{t-k+1},\ldots,x_{t-1})}{p(x_t\mid x_{t-k+1},\ldots,x_{t-1})\,p(x_{t-k}\mid x_{t-k+1},\ldots,x_{t-1})} \\
  =\;& \PMI\!\left(x_t;\,x_{t-k}\,\bigm|\,x_{t-k+1},\ldots,x_{t-1}\right),
\end{align*}
where the second equality uses \(p(X,Y\mid Z) = p(X\mid Y,Z)\,p(Y\mid Z)\), and the third uses the standard definition of conditional pointwise mutual information,
\[
\PMI(x;y\mid z)=\log\frac{p(x,y\mid z)}{p(x\mid z)\,p(y\mid z)}.
\]
Summing the difference \(\log p(x_t\mid x_{t-k},\ldots,x_{t-1}) - \log p(x_t\mid x_{t-k+1},\ldots,x_{t-1})\) over \(k=1,\ldots,t-1\) (the intermediate terms cancel with the convention that \(\log p(x_t\mid \emptyset) = \log p(x_t)\)) gives
\[
\log p(x_t\mid x_{<t})
= \log p(x_t)
+ \sum_{k=1}^{t-1} \PMI\!\left(x_t;\,x_{t-k}\,\bigm|\,x_{t-k+1},\ldots,x_{t-1}\right).
\]
The decomposition is exact, and identifies the unique additive split of \(\log p(x_t\mid x_{<t})\) into a unigram base and a sequence of higher-order PMI terms, each measuring the additional information that one further history token \(x_{t-k}\) provides about \(x_t\) given the more recent context.

\subsection{Analytic random-subspace expectations}
\label{app:haar-expectations}

Let \(d\geq 1\) be the representation dimension, let \(1\leq k\leq d\) be an integer, and fix a positive-semidefinite matrix \(\Gmat\in\mathbb{R}^{d\times d}\) with \(\tr(\Gmat)>0\).
Choose a \(k\)-dimensional subspace uniformly at random, meaning that its distribution is invariant under every orthogonal transformation of \(\mathbb{R}^d\) (the Haar distribution).
Let \(\Util\in\mathbb{R}^{d\times k}\) be any orthonormal basis for this subspace, so \(\Util^\top\Util=\Imat_k\), where \(\Imat_m\) denotes the \(m\times m\) identity matrix.
We show that the expected trace share is
\[
\mathbb{E}\!\left[
\frac{\tr(\Util^\top\Gmat\Util)}{\tr(\Gmat)}
\right]=\frac{k}{d},
\]
where the expectation is over the random subspace.

Let \(\Pmat=\Util\Util^\top\) be its orthogonal projector.
For every orthogonal matrix \(\Qmat\in\mathbb{R}^{d\times d}\), orthogonal invariance gives
\(\mathbb{E}[\Pmat]=\Qmat\mathbb{E}[\Pmat]\Qmat^\top\).
Flipping the sign of coordinate \(i\) negates the \((i,j)\) entry for \(i\neq j\), which forces all off-diagonal entries of \(\mathbb{E}[\Pmat]\) to vanish, and permuting coordinates makes all diagonal entries equal.
Thus \(\mathbb{E}[\Pmat]=c\Imat_d\) for some scalar \(c\).
Since \(\Pmat\) is a rank-\(k\) projector,
\(cd=\tr(\mathbb{E}[\Pmat])=\mathbb{E}[\tr(\Pmat)]=k\), so \(\mathbb{E}[\Pmat]=(k/d)\Imat_d\).
Cyclicity of the trace and linearity of expectation then give
\[
\mathbb{E}\!\left[\tr(\Util^\top\Gmat\Util)\right]
=\mathbb{E}[\tr(\Gmat\Pmat)]
=\tr\!\bigl(\Gmat\mathbb{E}[\Pmat]\bigr)
=\frac{k}{d}\tr(\Gmat).
\]
Dividing by the fixed positive quantity \(\tr(\Gmat)\) proves the claim.
In \cref{eq:optim-engagement}, we use this expected trace share \(k/d\) (with \(k=\rk(\Umat_{\ell,t})\)) as the baseline in the denominator inside the logarithm, so \(\oem_{\ell,t}=0\) means that the observed trace share equals the expected trace share of a uniformly random subspace of the same rank.
Note that this does not mean that a random subspace has expected \(\oem\) equal to zero: because \(\log_2\) is strictly concave, Jensen's inequality gives
\(\mathbb{E}\!\left[
  \log_2 \frac{d\,\tr(\Util^\top\Gmat\Util)}
               {k\,\tr(\Gmat)}
\right]\le
\log_2\mathbb{E}\!\left[
   \frac{d\,\tr(\Util^\top\Gmat\Util)}
               {k\,\tr(\Gmat)}
\right]=
\log_2 1 = 0\),
with strict inequality whenever the random trace share is non-constant.

\subsection{Noise scale of OE and the number of probe directions}
\label{app:sigma0}

In \Cref{app:haar-expectations}, we showed that on average, a uniformly random $k$-dimensional subspace captures a $k/d$ fraction of $\tr(\Gmat)$.
This average share corresponds to $\oem = 0$.
However, for any single random subspace, the actual captured share is not exactly $k/d$.
Because of random variation, a probe that learns nothing about the concept will still get a non-zero $\oem$ value by chance.
In this section, we derive the standard deviation $\sigma_0(k,\Gmat)$ of these random fluctuations.
This noise scale depends only on the AGOP matrix $\Gmat$ and the number of probe directions $k$.

For a single unit direction $\mathbf{u}\in\mathbb{R}^d$, we define its normalized AGOP gain relative to the random baseline as $g(\mathbf{u}) = \frac{d\,\mathbf{u}^\top \Gmat \mathbf{u}}{\tr(\Gmat)}$.
By \cref{eq:optim-engagement}, for a $k$-dimensional subspace with orthonormal basis $\{\mathbf{u}_1,\ldots,\mathbf{u}_k\}$, $\oem = \log_2 \bar{g}$, where $\bar{g} = \frac{1}{k}\sum_{j=1}^k g(\mathbf{u}_j)$ is the average gain across the $k$ directions.
When $\mathbf{u}$ is drawn uniformly from the unit sphere in $\mathbb{R}^d$, we can use the standard spherical moments $\mathbb{E}[u_i^4] = \frac{3}{d(d+2)}$ and $\mathbb{E}[u_i^2 u_j^2] = \frac{1}{d(d+2)}$ ($i\neq j$) to compute the mean and variance of $g$:
\begin{equation}
  \mathbb{E}[g] = 1,
  \qquad
  \operatorname{Var}(g)
  = \frac{2d}{d+2}\left(\frac{1}{\mathrm{PR}(\Gmat)} - \frac{1}{d}\right),
  \qquad
  \mathrm{PR}(\Gmat) = \frac{(\tr \Gmat)^2}{\tr(\Gmat^2)}.
  \label{eq:gain-variance}
\end{equation}
Here $\mathrm{PR}(\Gmat)$ is the \emph{participation ratio}: if $\Gmat$ has $r$ equal non-zero eigenvalues, then $\mathrm{PR}(\Gmat) = r$, so it measures the effective number of directions that carry the AGOP energy.
When the AGOP energy is concentrated in fewer directions, $\mathrm{PR}(\Gmat)$ is smaller, and the gain $g$ of a random direction fluctuates more ($\operatorname{Var}(g)\approx 2/\mathrm{PR}(\Gmat)$ when $d\gg\mathrm{PR}(\Gmat)$).
By contrast, when the AGOP energy is spread equally across all $d$ dimensions, $\mathrm{PR}(\Gmat) = d$ and $\operatorname{Var}(g) = 0$.

Averaging the gain over $k \ll d$ nearly independent orthonormal directions reduces the variance by a factor of $k$, giving $\operatorname{Var}(\bar{g}) \approx \operatorname{Var}(g)/k$.
Taking a first-order Taylor expansion of $\oem = \log_2 \bar{g}$ around $\mathbb{E}[\bar{g}] = 1$ gives $\oem \approx (\bar{g} - 1)/\ln 2$, which yields the null standard deviation of $\oem$:
\begin{equation}
  \sigma_0(k,\Gmat)
  \approx \frac{\sqrt{\operatorname{Var}(g)/k}}{\ln 2}.
  \label{eq:sigma0}
\end{equation}
Therefore, the random noise scale increases when $\Gmat$ is concentrated in fewer directions, and decreases with the number of probe directions at the rate $1/\sqrt{k}$.

\Cref{eq:sigma0} explains two empirical patterns in our results:
\begin{itemize}[leftmargin=1.5em,labelsep=0.3em,itemsep=2pt,topsep=2pt]
  \item \textbf{Varying $k$ within a task (\Cref{app:k-sensitivity}).}
  In Pythia-410M ($d=1024$), the median participation ratio of $\Gmat$ is about $85$.
  This gives $\sigma_0 \approx 0.05$ bits at $k=15$ (the full \pos probe), compared with $\sigma_0 \approx 0.21$ bits at $k=1$ (larger by a factor of $\sqrt{15}\approx 3.9$).
  This explains why reducing the \pos probe to $k=1$ in \Cref{tab:k-sweep} lowers the signal-to-noise ratio $\ce/\sigma_0$ and makes the $k=1$ curves visibly noisier.
  \item \textbf{Comparing across tasks (\Cref{subsec:ce-positive}).}
  At a given model, layer, and checkpoint, all tasks share the same AGOP matrix $\Gmat$, so their random noise scales $\sigma_0$ differ only through $k$.
  For example, n-gram regression probes use only $k=1$ direction, while the 16-class \pos probe uses $k=15$ directions.
  As a result, the random noise scale $\sigma_0$ for n-gram probes is $\sqrt{15}\approx 3.9$ times larger than for \pos.
  This higher noise level partly explains why the n-gram \cetext curves fluctuate more in \Cref{subsec:ce-positive}.
\end{itemize}

In practice, a control probe learns directions that have little relation to the target concept.
We therefore use the Haar calculation above as a reference for the seed-to-seed spread of control probes, and check this empirically below.
By contrast, real probes trained with different random seeds converge to almost the same subspace.
We verify this difference using the \pos probes from \Cref{app:independent-probes}.
Across the three seeds, the standard deviation of $\oem$ for the control probes is about $0.8\sigma_0$.
For the real probes, the standard deviation across seeds is only about $0.1\sigma_0$.
Therefore, the seed-to-seed noise in $\ce$ comes almost entirely from the control probes.
To reduce this noise, we average $\oem$ over three control seeds.

Finally, $\sigma_0$ describes the \emph{spread} (standard deviation) of the control $\oem$ distribution, not its \emph{mean}.
A control probe can still have a non-zero mean $\oem$ by memorizing word types.
We remove that mean shift by subtracting $\oem_{\text{ctrl}}$ in \cref{eq:concept-energy-def}, while we use $\sigma_0$ only to compare the noise level across settings such as different values of $k$ (\Cref{app:k-sensitivity}).

\section{Mean-centering and n-gram targets}
\label{app:centring}

\paragraph{Motivation.}
For all 12 tasks, we mean-center the probe inputs by subtracting the training-set mean activation $\bar{\hvec}$.
For the five n-gram regression tasks, we also center the regression targets by subtracting their training-set mean $\bar{y}$.
We mean-center the probe inputs for two reasons:
(a) transformer representations have a large common mean and a few dominant directions~\citep{mu2018allbutthetop,ethayarajh2019contextual,timkey2021allbark}; and
(b) recent probing work on LLMs also centers activations~\citep{burns2023discovering,marks-2024-the-geometry-of-truth}.
In our setting, there is another specific reason for centering both inputs and targets in the n-gram tasks: when the target mean $\bar{y}$ is large, a finite number of gradient-descent steps cannot fit this mean through the scalar bias alone, so the probe weights are forced to help fit the mean and form a dominant direction aligned with the input mean.

\paragraph{Gradient decomposition and the bias bottleneck.}
We write the activation of token $i$ as $\hvec_i = \bar{\hvec} + \tilde{\hvec}_i$, where $\bar{\hvec}$ is the training-set mean activation and $\sum_{i=1}^N \tilde{\hvec}_i = \mathbf{0}$.
For a linear regression probe $f(\hvec_i) = \wvec^\top \hvec_i + b_0$ with squared-error loss and residuals $r_i = f(\hvec_i) - y_i$, the gradients with respect to $\wvec$ and $b_0$ are
\begin{equation}
  \nabla_{\wvec} L
  = \frac{1}{N}\sum_{i=1}^N r_i\,\tilde{\hvec}_i + \bar{r}\,\bar{\hvec},
  \qquad
  \frac{\partial L}{\partial b_0} = \bar{r},
  \label{eq:surrogate-bias-grad}
\end{equation}
where $\bar{r} = \frac{1}{N}\sum_{i=1}^N r_i$ is the mean residual.
At initialization ($\wvec\approx\mathbf{0}$, $b_0\approx 0$), the mean residual is $\bar{r}\approx -\bar{y}$.
For the n-gram tasks, the target mean $|\bar{y}|$ is often large (for example, $\bar{y} = -6.93$ for \unigramlp, $-3.54$ for \bigramlp, and $+2.91$ for \bigrampmi; \Cref{tab:ngram-fit}).
However, with slowly varying, constant-sign bias gradients, Adam typically changes the scalar bias $b_0$ by roughly $\eta$ per step ($\eta = 10^{-3}$).
Over $T$ training steps (10 epochs), $\eta T$ therefore sets the characteristic scale of $|b_0|$, and our training budget is only $\eta T \in [0.66, 4.25]$ (\Cref{tab:ngram-fit}).
Consequently, when $|\bar{y}| > \eta T$ (which holds for \unigramlp, \bigramlp, and \bigrampmi),  this training budget is too small for the bias $b_0$ to reach $\bar{y}$.
As a result, the mean residual $\bar{r}$ stays large throughout training.

\paragraph{Two failure modes when targets are uncentered.}
When $|\bar{y}| > \eta T$ and the targets are not centered ($\bar{y}\neq 0$), a large mean residual $\bar{r}$ in \cref{eq:surrogate-bias-grad} causes one of two problems, depending on whether the inputs are centered:
\begin{itemize}[leftmargin=1.5em,labelsep=0.3em,itemsep=2pt,topsep=2pt]
  \item \textbf{Uncentered inputs and uncentered targets ($\bar{\hvec}\neq\mathbf{0}$, $\bar{y}\neq 0$): a surrogate bias inflates control $\oem$.}
  Because $\bar{r}$ keeps the sign of $-\bar{y}$, the term $\bar{r}\bar{\hvec}$ in \cref{eq:surrogate-bias-grad} pushes $\wvec$ along a mean-related direction $\mathbf{v}$ at every step.
  This direction $\mathbf{v}$ depends on the input mean $\bar{\hvec}$ and the sign of $\bar{y}$, so the dot product $\wvec^\top\bar{\hvec}$ has the same sign as $\bar{y}$.
  In effect, the probe uses $\wvec^\top\bar{\hvec}$ to help fit the target mean $\bar{y}$ because the scalar bias $b_0$ moves too slowly.
  We call this direction a \emph{surrogate bias}.
  For a real probe, the signal term $\frac{1}{N}\sum_i r_i\tilde{\hvec}_i$ also pulls $\wvec$ toward task-relevant directions.
  For a control probe, however, the signal term is mostly random noise.
  The control probe's weight vector therefore points almost entirely along the surrogate-bias direction $\hat{\mathbf{v}} = \frac{\mathbf{v}}{\|\mathbf{v}\|_2}$.
  Its $\oem$ is then approximately $\log_2(\frac{d\,\hat{\mathbf{v}}^\top \Gmat \hat{\mathbf{v}}}{\tr\Gmat})$.
  Whenever $\Gmat$ has high energy along $\hat{\mathbf{v}}$, the control probe's $\oem$ becomes artificially large.
  \item \textbf{Centered inputs and uncentered targets ($\bar{\hvec}=\mathbf{0}$, $\bar{y}\neq 0$): a constant prediction offset makes $R^2$ negative.}
  Mean-centering the inputs sets $\bar{\hvec}=\mathbf{0}$ and removes the $\bar{r}\bar{\hvec}$ term.
  Because the centered inputs have zero mean, the weights $\wvec$ can no longer help fit the target mean $\bar{y}$, leaving the scalar bias $b_0$ to fit $\bar{y}$ on its own.
  For the three tasks with $|\bar{y}| > \eta T$, $b_0$ fails to reach $\bar{y}$ within $T$ training steps, so the predictions stay far from the true target mean.
  Although the Pearson correlation $r$ between the held-out predictions and the targets remains nearly unchanged, this large gap in the mean makes the held-out $R^2$ negative (Uncentered $y$ columns of \Cref{tab:ngram-fit}).
\end{itemize}

\paragraph{Centering both inputs and targets.}
To avoid both problems, we subtract the training-set mean $\bar{y}$ (computed in float64) from the targets of every n-gram probe (real or control), and we subtract the same training-set mean from the test-set targets.
Once the targets are centered, their mean is zero from the start of training ($\bar{y}=0$, so $\bar{r}\approx 0$).
As a result, neither the bias $b_0$ nor the weights $\wvec$ need to fit a large target mean, and the $\bar{r}\bar{\hvec}$ term in \cref{eq:surrogate-bias-grad} disappears even when the inputs are not centered.
As shown in \Cref{tab:ngram-fit}, centering the targets restores positive held-out $R^2$ across all tasks.
By contrast, leaving $y$ uncentered makes held-out $R^2$ negative ($-1.19$ to $-2.99$) for the three tasks with $|\bar{y}| > \eta T$, while the two trigram tasks ($|\bar{y}| < \eta T = 4.25$) are unaffected.
Classification tasks do not have a scalar target offset to fit; we mean-center their inputs as well so that all 12 tasks use the same input preprocessing.

\begin{table}[!htbp]
  \centering\small
  \caption{
    Held-out fit of the real n-gram probes with mean-centered inputs, for centered and uncentered targets $y$.
    $r$: Pearson correlation between the probe's held-out predictions and the targets; $R^2$: held-out coefficient of determination.
    Entries are medians over layers and checkpoints 2k--143k of the three-seed mean.
    $\bar{y}$: training-set target mean; $\eta T$: learning rate ($10^{-3}$) times the number of optimizer steps over 10 epochs.
    When $|\bar{y}| > \eta T$ (\unigramlp, \bigramlp, and \bigrampmi), uncentered targets leave $r$ nearly unchanged but make the held-out $R^2$ negative.
  }
  \begin{tabular}{lrrrrrrrrrr}
\toprule
& & & \multicolumn{4}{c}{Pythia-410M} & \multicolumn{4}{c}{Pythia-1B} \\
\cmidrule(lr){4-7}\cmidrule(l){8-11}
& & & \multicolumn{2}{c}{Centered $y$} & \multicolumn{2}{c}{Uncentered $y$} & \multicolumn{2}{c}{Centered $y$} & \multicolumn{2}{c}{Uncentered $y$} \\
\cmidrule(lr){4-5}\cmidrule(lr){6-7}\cmidrule(lr){8-9}\cmidrule(l){10-11}
Task & $\bar y$ & $\eta T$ & $r$ & $R^2$ & $r$ & $R^2$ & $r$ & $R^2$ & $r$ & $R^2$ \\
\midrule
Unigram LP & $-6.93$ & 1.46 & $0.93$ & $0.83$ & $0.91$ & $-2.99$ & $0.94$ & $0.86$ & $0.93$ & $-2.87$ \\
Bigram LP & $-3.54$ & 0.66 & $0.85$ & $0.71$ & $0.85$ & $-1.38$ & $0.86$ & $0.73$ & $0.86$ & $-1.19$ \\
Bigram PMI & $+2.91$ & 0.66 & $0.68$ & $0.36$ & $0.68$ & $-2.28$ & $0.72$ & $0.42$ & $0.72$ & $-2.47$ \\
Trigram LP & $-2.29$ & 4.25 & $0.82$ & $0.64$ & $0.82$ & $0.64$ & $0.84$ & $0.68$ & $0.84$ & $0.68$ \\
Trigram PMI & $+0.99$ & 4.25 & $0.42$ & $0.05$ & $0.42$ & $0.05$ & $0.45$ & $0.04$ & $0.45$ & $0.05$ \\
\bottomrule
\end{tabular}

  \label{tab:ngram-fit}
\end{table}

\paragraph{$2\times 2$ comparison of input and target centering.}
\Cref{tab:centring-grid} tests all four combinations of input centering and target centering on a $5\times 5$ grid in Pythia-410M (checkpoints 128, 256, 512, 1{,}000, and 143{,}000 $\times$ layers 5, 9, 13, 17, and 21).
The three tasks with $|\bar{y}| > \eta T$ match the mechanism above:
(a) when neither inputs nor targets are centered, the surrogate bias fits the target offset so real-probe $R^2$ is positive, but control $\oem$ is inflated ($+0.41$ to $+0.65$);
(b) when only inputs are centered, the surrogate bias is removed and real-probe $R^2$ becomes negative in $100\%$ of cells ($-1.57$ to $-2.75$); and
(c) once targets are centered, real-probe $R^2$ is positive and control $\oem$ drops near zero ($-0.06$ to $+0.14$) whether or not inputs are centered.
For the two trigram tasks, $|\bar{y}| < \eta T$, so all four combinations give similar results.

\begin{table}[!htbp]
  \centering\small
  \setlength{\tabcolsep}{4.5pt}
  \caption{
    $2\times 2$ ablation of input mean-centering and target centering for the five n-gram tasks in Pythia-410M, on a grid of 5 checkpoints (128, 256, 512, 1k, 143k) $\times$ 5 layers.
    Each entry is the median over the 25 cells of the real-probe \(R^2\), the control \oetext, or \cetext; in parentheses, the share of cells with \(R^2<0\) (real-probe \(R^2\)) or with \(\ce>0\) (\cetext).
    $\bar{y}$: training-set target mean; $\eta T$: learning rate ($10^{-3}$) times the number of optimizer steps over 10 epochs.
  }
  \begin{tabular}{lrrrrrr}
\toprule
& & & \multicolumn{2}{c}{Uncentered targets} & \multicolumn{2}{c}{Centered targets} \\
\cmidrule(lr){4-5}\cmidrule(lr){6-7}
Task & $\bar y$ & $\eta T$ & Raw inputs & Centered inputs & Raw inputs & Centered inputs \\
\midrule
\multicolumn{7}{l}{\emph{Real-probe $R^2$}} \\
Unigram LP & $-6.93$ & 1.46 & $+0.66$ (0\%) & $-2.75$ (100\%) & $+0.53$ (0\%) & $+0.51$ (0\%) \\
Bigram LP & $-3.54$ & 0.66 & $+0.63$ (0\%) & $-1.57$ (100\%) & $+0.65$ (0\%) & $+0.65$ (0\%) \\
Bigram PMI & $+2.91$ & 0.66 & $+0.27$ (0\%) & $-2.03$ (100\%) & $+0.27$ (0\%) & $+0.27$ (0\%) \\
Trigram LP & $-2.29$ & 4.25 & $+0.48$ (0\%) & $+0.47$ (0\%) & $+0.48$ (0\%) & $+0.48$ (0\%) \\
Trigram PMI & $+0.99$ & 4.25 & $-0.02$ (60\%) & $-0.01$ (64\%) & $-0.01$ (52\%) & $-0.01$ (64\%) \\
\midrule
\multicolumn{7}{l}{\emph{Control \oetext}} \\
Unigram LP & $-6.93$ & 1.46 & $+0.41$ & $+0.26$ & $-0.02$ & $-0.06$ \\
Bigram LP & $-3.54$ & 0.66 & $+0.65$ & $+0.31$ & $+0.14$ & $+0.13$ \\
Bigram PMI & $+2.91$ & 0.66 & $+0.55$ & $+0.21$ & $+0.14$ & $+0.09$ \\
Trigram LP & $-2.29$ & 4.25 & $+0.05$ & $-0.15$ & $-0.13$ & $-0.19$ \\
Trigram PMI & $+0.99$ & 4.25 & $-0.45$ & $-0.51$ & $-0.51$ & $-0.51$ \\
\midrule
\multicolumn{7}{l}{\emph{\cetext}} \\
Unigram LP & $-6.93$ & 1.46 & $+0.53$ (80\%) & $+0.98$ (100\%) & $+0.72$ (84\%) & $+0.70$ (88\%) \\
Bigram LP & $-3.54$ & 0.66 & $+0.16$ (76\%) & $+0.50$ (96\%) & $+0.38$ (88\%) & $+0.41$ (88\%) \\
Bigram PMI & $+2.91$ & 0.66 & $+0.32$ (80\%) & $+0.51$ (84\%) & $+0.55$ (84\%) & $+0.51$ (84\%) \\
Trigram LP & $-2.29$ & 4.25 & $+0.01$ (52\%) & $+0.11$ (72\%) & $+0.12$ (60\%) & $+0.12$ (68\%) \\
Trigram PMI & $+0.99$ & 4.25 & $+0.40$ (84\%) & $+0.25$ (84\%) & $+0.39$ (88\%) & $+0.25$ (84\%) \\
\bottomrule
\end{tabular}

  \label{tab:centring-grid}
\end{table}

\begin{figure}[!tbp]
  \centering
  \includegraphics[width=\linewidth]{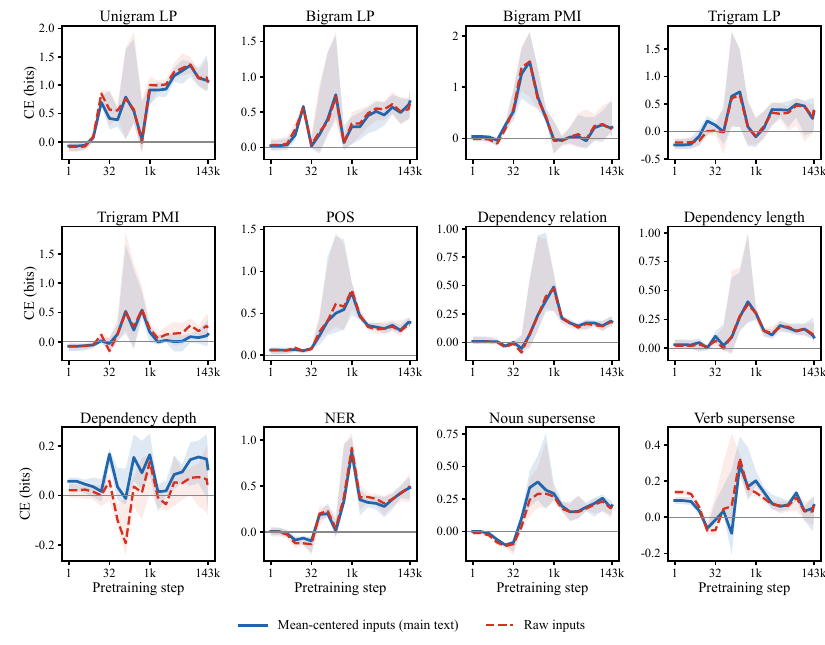}
  \caption{
    \cetext over pretraining for the 12 tasks in Pythia-410M with mean-centered probe inputs (blue, solid; as in the main text) and raw inputs (red, dashed); n-gram targets are centered in both.
    Lines and bands: median and 25th--75th percentiles across the 24 layers.
  }
  \label{fig:raw-vs-centred-410m}
\end{figure}

\paragraph{Effect of input centering on the main conclusions.}
Finally, \Cref{fig:raw-vs-centred-410m} and \Cref{tab:raw-vs-centred-410m} compare the full trajectories of Pythia-410M (19 checkpoints $\times$ 24 layers) across all 12 tasks with and without input mean-centering (with n-gram targets centered in both cases).
For most tasks, the two \cetext curves almost overlap, with Spearman correlations of $\rho \in [0.81, 0.99]$ across (layer, checkpoint) cells from step 64 onward.
The largest difference is in \depdepth, which also has the smallest \cetext values overall.
Across all 12 tasks, the task-level timing-order counts are \texttt{11/1/0} in both settings.

\begin{table}[!tbp]
  \centering\small
  \setlength{\tabcolsep}{2.5pt}
  \caption{
    Mean-centered versus raw probe inputs in Pythia-410M.
    Step columns: median \cetext (bits) over the (layer, checkpoint) cells at steps 1--32 and 64--143k, with the share of cells with \(\ce>0\) in parentheses.
    $\rho$: Spearman correlation between the two \cetext values over the cells from step 64 onward.
    Lag: \(\ln(t^E/t^D)\) of the task-level events (\cref{eq:aggregated-event-time}), with decodability measured by selectivity; E: energy event earlier; =: the two events coincide.
  }
  \begin{tabular}{lrrrrrrr}
\toprule
 & \multicolumn{2}{c}{1--32} & \multicolumn{2}{c}{64--143k} & & \multicolumn{2}{c}{Lag} \\
Task & Centered & Raw & Centered & Raw & $\rho$ & Centered & Raw \\
\midrule
Unigram LP & $+0.03$ (56\%) & $+0.06$ (56\%) & $+1.01$ (95\%) & $+1.08$ (95\%) & 0.91 & $-1.07$ E & $-0.44$ E \\
Bigram LP & $+0.10$ (74\%) & $+0.12$ (78\%) & $+0.46$ (90\%) & $+0.48$ (92\%) & 0.97 & $-3.98$ E & $-2.79$ E \\
Bigram PMI & $+0.06$ (71\%) & $+0.03$ (60\%) & $+0.34$ (72\%) & $+0.36$ (74\%) & 0.99 & $-1.92$ E & $-1.69$ E \\
Trigram LP & $-0.11$ (31\%) & $-0.12$ (24\%) & $+0.30$ (76\%) & $+0.28$ (72\%) & 0.98 & $-1.62$ E & $-1.56$ E \\
Trigram PMI & $-0.05$ (33\%) & $-0.07$ (28\%) & $+0.11$ (69\%) & $+0.22$ (82\%) & 0.94 & $-0.92$ E & $-1.06$ E \\
POS & $+0.07$ (80\%) & $+0.06$ (80\%) & $+0.38$ (100\%) & $+0.38$ (100\%) & 0.95 & $-0.03$ = & $-0.10$ = \\
Dependency relation & $0.00$ (49\%) & $-0.01$ (46\%) & $+0.18$ (91\%) & $+0.17$ (91\%) & 0.98 & $-0.43$ E & $-0.36$ E \\
Dependency length & $+0.03$ (74\%) & $+0.03$ (69\%) & $+0.16$ (93\%) & $+0.17$ (92\%) & 0.96 & $-1.90$ E & $-0.97$ E \\
Dependency depth & $+0.05$ (78\%) & $+0.02$ (65\%) & $+0.08$ (76\%) & $+0.02$ (54\%) & 0.81 & $-1.15$ E & $-0.97$ E \\
NER & $-0.05$ (26\%) & $-0.07$ (24\%) & $+0.34$ (96\%) & $+0.36$ (96\%) & 0.98 & $-0.61$ E & $-0.51$ E \\
Noun supersense & $-0.04$ (20\%) & $-0.04$ (7\%) & $+0.20$ (98\%) & $+0.19$ (96\%) & 0.94 & $-0.58$ E & $-0.46$ E \\
Verb supersense & $+0.07$ (69\%) & $+0.11$ (74\%) & $+0.10$ (84\%) & $+0.09$ (85\%) & 0.90 & $-4.07$ E & $-3.33$ E \\
\bottomrule
\end{tabular}

  \label{tab:raw-vs-centred-410m}
\end{table}

\section{Complete results}
\label{app:complete-results}

\subsection{Complete results of \cetext}
\label{app:subsec:timecourse-curve-heatmap}

\begin{figure}[!tbp]
  \centering
  \includegraphics[width=\linewidth]{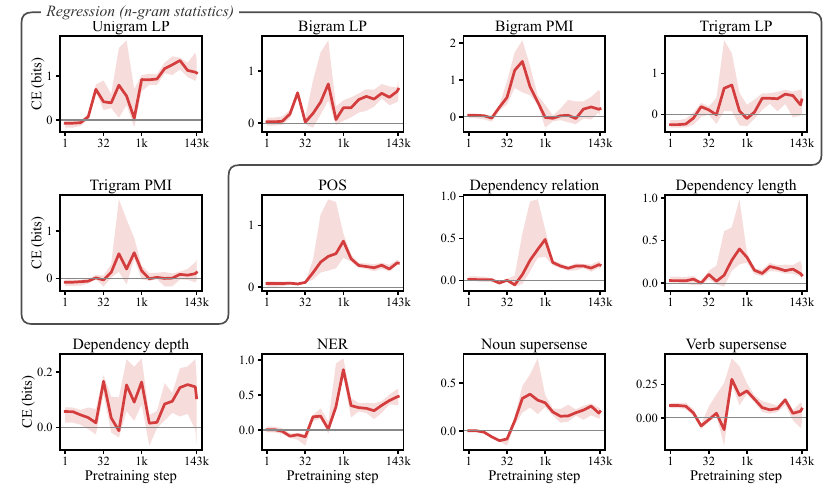}
  \caption{
    \cetext over pretraining for the 12 tasks in Pythia-410M.
    Lines and bands: median and 25th--75th percentiles across the 24 layers.
    The outlined panels are the five regression (n-gram) tasks; the other seven are classification tasks.
    Y-axes differ across panels.
    The share of cells with \(\ce>0\) is in \Cref{tab:ce-positive-share}.
  }
  \label{fig:ce-timecourse-410m}
\end{figure}

\begin{figure}[!tbp]
  \centering
  \includegraphics[width=\linewidth]{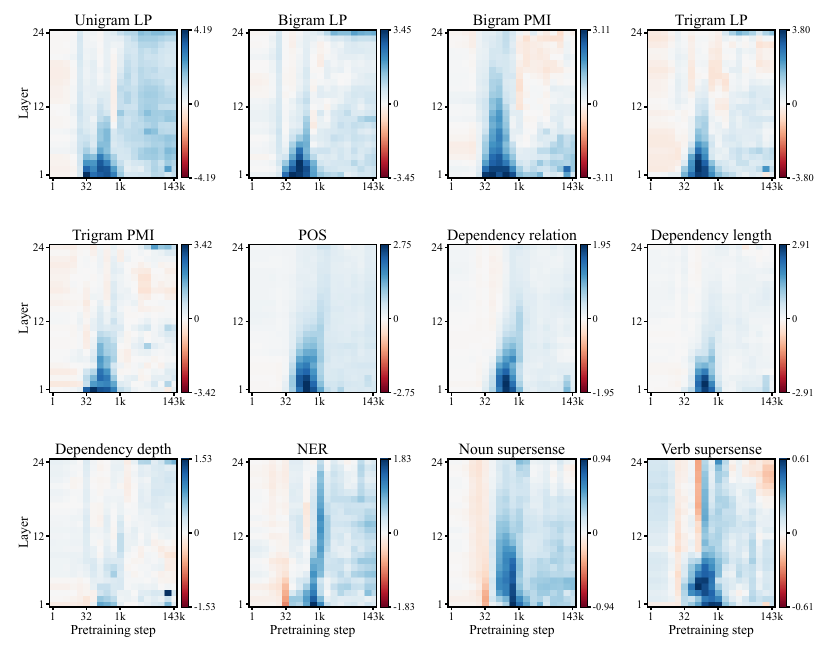}
  \caption{
    Per-layer \cetext over pretraining for the 12 tasks in Pythia-410M.
    Rows: layers, numbered from 1 (shallowest); columns: the 19 checkpoints, equally spaced. Blue: \(\ce>0\); red: \(\ce<0\); darker cells have larger \(|\ce|\).
    Each panel has its own color scale, symmetric about 0, so colors are not comparable across panels.
  }
  \label{fig:ce-heatmap-410m}
\end{figure}

\Cref{fig:ce-timecourse-410m} shows how \(\ce\) changes over pretraining in Pythia-410M.
Its overall pattern is similar to that of Pythia-1B in \Cref{fig:ce-timecourse-1b}: among the classification tasks, all except \depdepth exhibit a clear peak around steps 256--1k.
Among the n-gram tasks, \unigramlp, \bigramlp, and \trigramlp continue to rise in later training, while \bigrampmi shows a clear peak and remains low in later stages.
\trigrampmi fluctuates at a low level overall, although 8 of its 10 shallowest layers reach their highest peak at step 128 (\Cref{fig:ce-heatmap-410m}), which slightly elevates the median curve and produces a wider interquartile band across steps 64--1k.

\begin{figure}[!tbp]
  \centering
  \includegraphics[width=\linewidth]{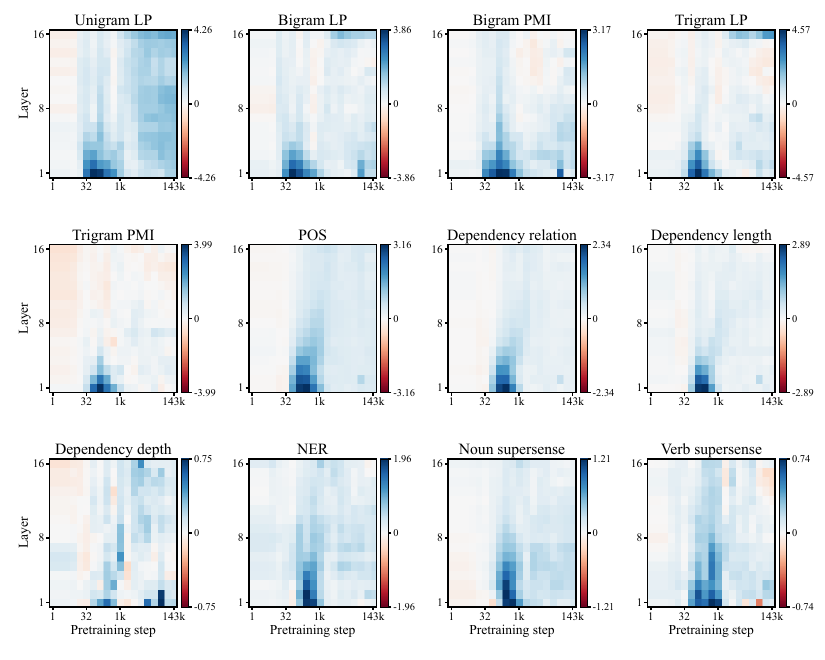}
  \caption{Per-layer \cetext over pretraining for the 12 tasks in Pythia-1B, drawn as in \Cref{fig:ce-heatmap-410m}.}
  \label{fig:ce-heatmap-1b}
\end{figure}

While \Cref{fig:ce-timecourse-410m,fig:ce-timecourse-1b} plot layer medians and quartiles, \cetext in fact exhibits systematic pattern differences across shallow and deep layers.
\Cref{fig:ce-heatmap-410m,fig:ce-heatmap-1b} show per-layer heatmaps of \(\ce\) over time for both models: almost all tasks exhibit relatively higher \(\ce\) in shallow layers, and most tasks display a clean single peak in shallow layers.
In particular, \unigramlp shows high \ce~during middle and late training across layers, especially in deep layers, which have almost no early peak and instead rise in middle-to-late training and remain high, whereas \depdepth stays generally low throughout pretraining in both models, with well-formed peaks harder to observe.

\subsection{Comparing \oetext and \cetext throughout pretraining}
\label{app:subsec:oe-control-ce}

\begin{figure}[!tbp]
  \centering
  \includegraphics[width=\linewidth]{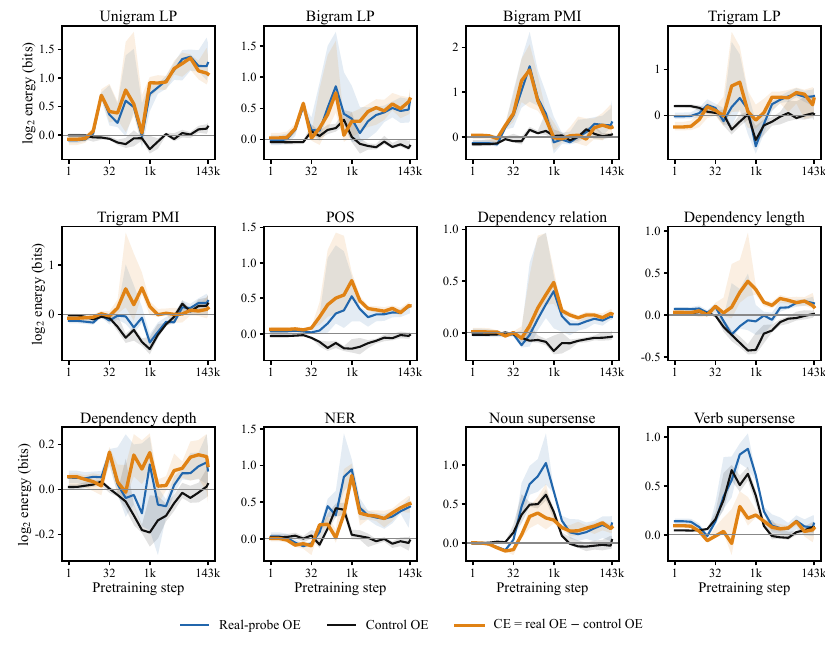}
  \caption{
    \(\oem\) of the real task (blue) and of the control task (black), and their difference \(\ce=\oem_{\text{real}}-\oem_{\text{ctrl}}\) (orange), over pretraining for the 12 tasks in Pythia-410M.
    Lines and bands: median and 25th--75th percentiles across the 24 layers. Y-axes differ across panels.
  }
  \label{fig:oe_control_ce_410m}
\end{figure}

\begin{figure}[!tbp]
  \centering
  \includegraphics[width=\linewidth]{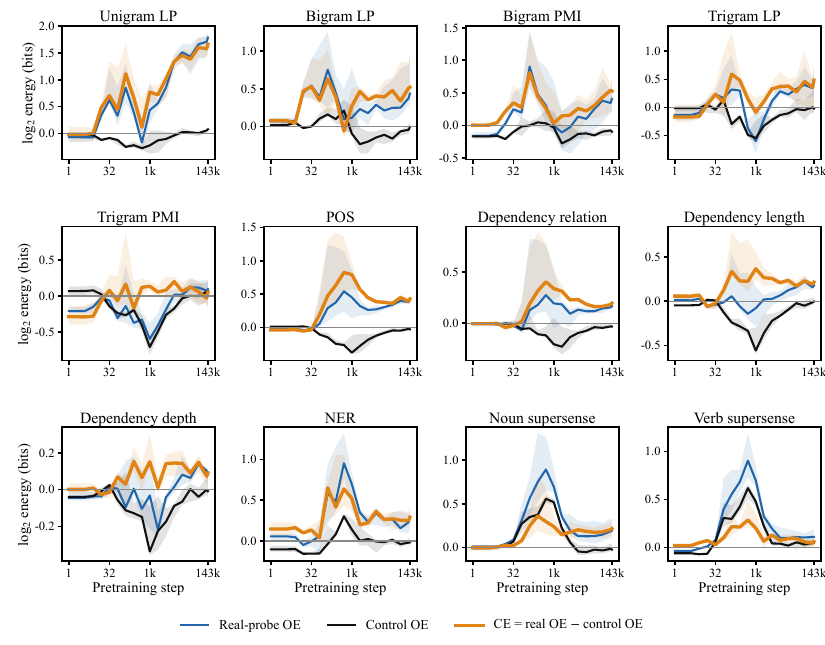}
  \caption{\(\oem\) of the real and control tasks and \cetext over pretraining for the 12 tasks in Pythia-1B, drawn as in \Cref{fig:oe_control_ce_410m} (median over the 16 layers).}
  \label{fig:oe_control_ce_1b}
\end{figure}

In \Cref{fig:oe_control_ce_410m,fig:oe_control_ce_1b}, we compare the \(\ce\) curves of Pythia-410M and -1B with the separate \(\oem\) curves of the real and control tasks.
The patterns are very similar across the two models.
Specifically,
\begin{itemize}[
    leftmargin=1.5em,
    labelsep=0.3em,
    itemsep=0pt
]
    \item In the cleanest pattern, \(\oem\) is positive for the real task and negative for the control task, which holds for \unigramlp, \bigrampmi, \pos, and \deprel.
    \item For \bigramlp, \ner, \ssnoun, and \ssverb, both the real and control tasks have positive \(\oem\), but the real task's \(\oem\) is almost always larger. 
    For these tasks, examining the real task's \(\oem\) alone may overestimate concept-specific enrichment.
    \item In a third pattern, the real task's \(\oem\) is small, fluctuating around 0 or even becoming negative, as in \trigramlp, \trigrampmi, \deplen, and \depdepth. 
    Specifically, for \trigramlp, the real and control \(\oem\) curves stay close and are both negative up to around step 1k, after which the real task's \(\oem\) gradually rises and turns positive while the control task's \(\oem\) remains below 0 and eventually fluctuates near 0; for \trigrampmi, the gap between the real and control \(\oem\) is small, leaving \(\ce\) fluctuating just above 0 and occasionally dipping below 0; for \deplen and \depdepth, the control task's \(\oem\) is even lower than that of the real task, so \(\ce\) still remains positive overall during pretraining.
    For this pattern, examining the real task's \(\oem\) alone may underestimate concept-specific enrichment.
\end{itemize}

\clearpage

\section{Event timing: trajectories, summary sensitivity, and qualifying peaks}
\label{app:probing-timing-analyses}

\subsection{Trajectories of decodability and \cetext with event times}
\label{app:subsec:trajectories}

\begin{figure}[!tbp]
  \centering
  \includegraphics[width=\linewidth]{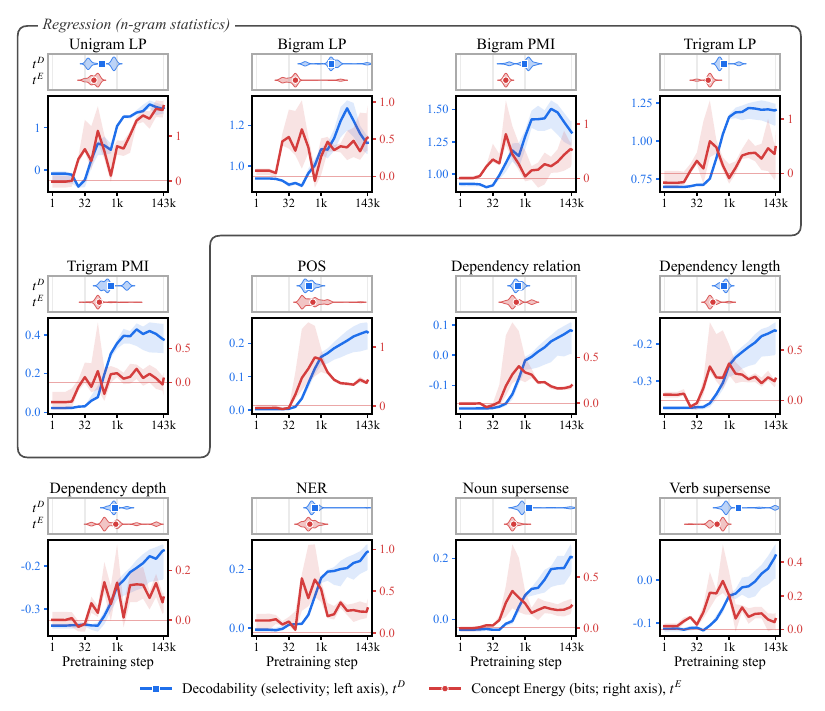}
  \caption{Decodability (selectivity; blue, left axis) and \cetext (bits; red, right axis) over pretraining for the 12 tasks in Pythia-1B.
  Lines and bands: median and 25th--75th percentiles across the 16 layers, each layer averaged over probe seeds; the two y-axes are scaled separately and differ across panels.
  The strip above each panel marks $t^D$ and $t^E$ (\cref{eq:aggregated-event-time}), the coordinates of the task's point in \Cref{fig:probe-decode-diagonal}; violins show the $w_\ell$-weighted distribution of the layer events, as in \Cref{fig:sva-decode-energy-behavior}.
  Events are computed per layer and then aggregated, so they need not coincide with features of the plotted medians.
  The outlined panels are the five regression (n-gram) tasks; the other seven are classification tasks.}
  \label{fig:traj-events-1b}
\end{figure}

\begin{figure}[!tbp]
  \centering
  \includegraphics[width=\linewidth]{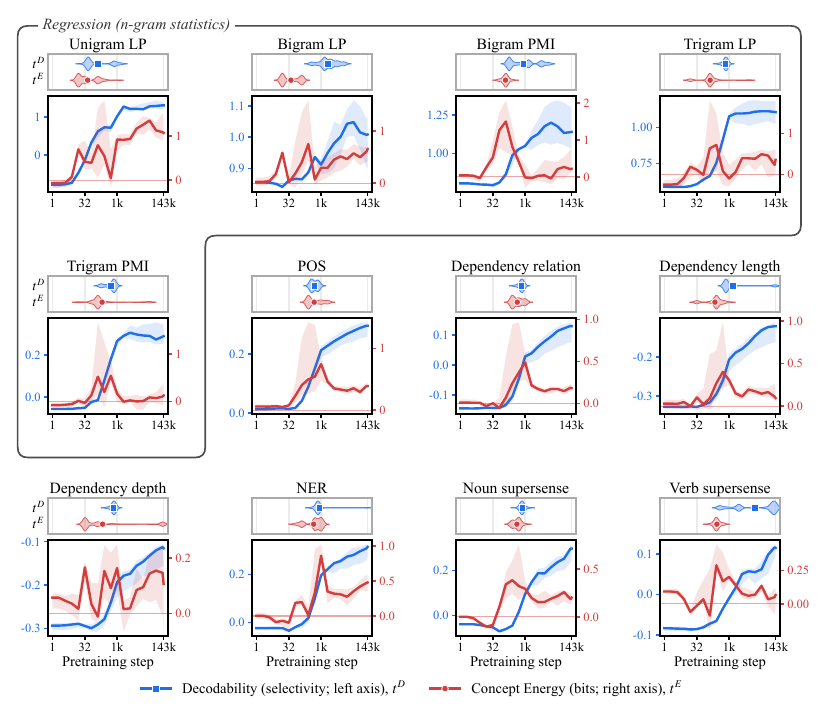}
  \caption{As \Cref{fig:traj-events-1b}, for Pythia-410M: decodability (selectivity; blue, left axis) and \cetext (bits; red, right axis), with the median and 25th--75th percentiles across the 24 layers, and the event strip of each task.}
  \label{fig:traj-events-410m}
\end{figure}

\Cref{fig:traj-events-1b,fig:traj-events-410m} show how decodability and \cetext change over pretraining in Pythia-1B and -410M, respectively, and mark the cross-layer aggregated energy events and decodability transitions for each task.
Consistent with \cref{subsec:timing-diagonal}, $t^E$ precedes $t^D$ in most tasks; the exceptions are \pos in Pythia-1B, where $t^D$ is earlier, as well as \deprel and \depdepth in Pythia-1B and \pos in Pythia-410M, where the two events coincide.
In addition, for both Pythia-410M and -1B, selectivity exceeds 1 for some regression tasks and stays negative for \deplen and \depdepth (\Cref{app:selectivity}).

\subsection{Temporal ordering and summary choice sensitivity}
\label{app:subsec:temporal-summary-order-sensitivity}

The task-level probing event-time plot (\Cref{fig:probe-decode-diagonal}) aggregates layer-wise events using geometric means weighted by $w_\ell=\max_i[\ce_{\ell,t_i}]_+$ (\cref{eq:aggregated-event-time}), which can obscure differences in temporal ordering across layers.
Moreover, changing the method to identify energy events can result in different temporal ordering.
In this section, we compare different temporal summaries for \cetext and provide layer-wise statistics for how the summary choice can change the temporal ordering.

\textbf{Event lag.}\quad
For layer \(\ell\) in which both \cetext event and decodability transition defined in \cref{subsec:event-timing} exist, we define the signed \textbf{event lag} as
\begin{equation}
  \Delta_{\ell}
  =
  \ln t^{E}_{\ell}
  -
  \ln t^{D}_{\ell}
  =\ln\left(\frac{t^{E}_{\ell}}{t^{D}_{\ell}}\right),
  \label{eq:event-lag}
\end{equation}
in natural-log training-step units.
Negative values indicate that energy events precede the decodability transitions; positive values indicate the reverse.

\textbf{Temporal summary.}\quad
We distinguish three ways to summarize a \cetext trajectory into an energy event as follows:
\begin{itemize}[leftmargin=2em,topsep=2pt,itemsep=2pt]
    \item $\tau_{\text{1st}}$: the training step when an energy trajectory reaches its earliest qualifying peak.
    A step qualifies as a peak when it meets the conditions described in \cref{app:peak-rule}.
    This is exactly the $t^E_\ell$ in \cref{subsec:event-timing};
    \item $\tau_{\max}$: the training step with maximum energy values throughout an energy trajectory.
    It is not constrained by the rules in \cref{app:peak-rule};
    \item $\tau_{50}$: the training step at which cumulative positive \cetext area, integrated over log training step, reaches half of its total. Let $x_i=\ln t_i$ and $z_i=\ce_{t_i}$. We linearly interpolate the signed values $z_i$ between adjacent checkpoints to obtain $z(x)$ and define
    \begin{align}
    A &=\int_{x_1}^{x_N}[z(u)]_+\,du,
    \notag\\
    x_{50} &=\min\left\{x\in[x_1,x_N]:\int_{x_1}^{x}[z(u)]_+\,du\ge A/2\right\},
    \notag\\
    \tau_{50} &=\exp(x_{50}).
    \label{eq:definition-tau50}
    \end{align}%
\end{itemize}

In \cref{tab:event-summaries}, we report the distribution of timing statistics across layers for each temporal summary and task.
In the ``Layers'' columns we report the numbers of layers at which the energy event occurs earlier than, coincides with, or occurs later than the decodability event in the form of \texttt{earlier/same/later}.\footnote{%
    Here it is the cross-layer timing-order count, not the cross-task one used in \cref{subsec:timing-diagonal}.
}
It is evident that the $\tau_{\max}$ and $\tau_{50}$ generally identify later steps, resulting in energy events on more layers occurring \emph{after} the decodability transitions. 
Specifically, the number of layers with a later energy event rises from 28 ($\tau_{\text{1st}}$) to 81 ($\tau_{\max}$) to 139 ($\tau_{50}$) of 288 in Pythia-410M, and from 40 to 77 to 92 of 189 in Pythia-1B.

\begin{figure}[!tbp]
  \centering
  \includegraphics[width=\linewidth]{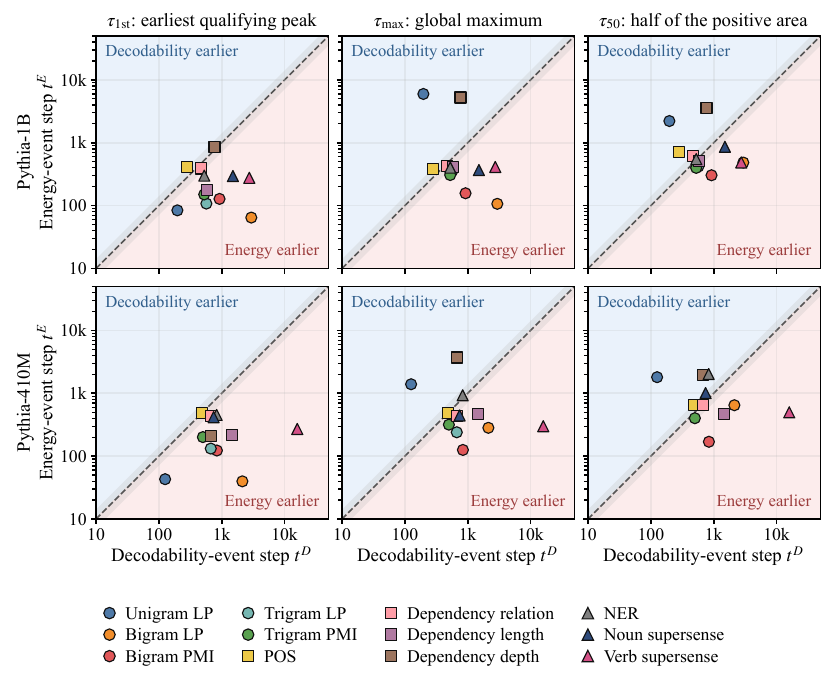}
  \caption{Task-level energy-event step against decodability-event step under the three definitions of the energy event (columns), for Pythia-1B (top row) and Pythia-410M (bottom row).
  Each point is one task, aggregated over layers by \cref{eq:aggregated-event-time}; points below the diagonal have the energy event earlier, and the gray band marks $\frac{\sqrt{2}}{2} <\frac{t^E}{t^D}<\sqrt{2}$.
  The first column shows the same task points as \Cref{fig:probe-decode-diagonal}.}
  \label{fig:event-summaries}
\end{figure}

At the task level (\Cref{fig:event-summaries}), when the energy event is defined as $\tau_{\max}$, the timing-order counts are \texttt{8/2/2} (410M) and \texttt{7/3/2} (1B); when defined as $\tau_{50}$, they are \texttt{4/5/3} (410M) and \texttt{4/5/3} (1B).
\Cref{fig:event-summaries} shows the per-task temporal ordering for Pythia-410M and -1B under the different temporal summaries.
When switching to $\tau_{\max}$, \unigramlp and \depdepth shift to having an earlier decodability transition for both Pythia-410M and -1B, because $\tau_{\text{1st}}<\tau_{\max}$ for both tasks and $\tau_{\max}$ falls after $t^D$; conversely, \pos in Pythia-1B shifts from an earlier decodability transition to coinciding.
When switching to $\tau_{50}$, the tasks with an earlier decodability transition are \unigramlp, \pos, and \depdepth in Pythia-1B, and \unigramlp, \depdepth, and \ner in Pythia-410M.
This is because $\tau_{50}$ generally occurs later: for many tasks, the \cetext curve stays close to 0 early on and remains positive for a long period after passing the earliest peak.
Overall, whether an energy event necessarily precedes decodability transitions depends on the definition of the energy event.

\begin{table}[t]\centering\small
    \caption{Layer-level timing order under three definitions of the energy event.
    For each task and summary, Layers gives the number of layers in which the energy event is earlier than, coincides with, or is later than the decodability event (\texttt{earlier/same/later}; the two coincide when $|\Delta_\ell|<\ln 2/2$), followed by the median and quartiles of the signed lag $\Delta_\ell$ (\cref{eq:event-lag}; negative: energy event earlier), with every layer counted once.
    All three summaries use the layers that enter \cref{eq:aggregated-event-time}: 24 per task in Pythia-410M and 16 in Pythia-1B, except 14 for \unigramlp and 15 for \trigramlp, which have layers without a qualifying peak.
    \emph{All tasks} pools the layers of the 12 tasks.}
    \resizebox{\linewidth}{!}{\begin{tabular}{@{}llrr@{\,}lrr@{\,}l@{}}
\toprule
& & \multicolumn{3}{c}{Pythia-1B} & \multicolumn{3}{c}{Pythia-410M} \\
\cmidrule(lr){3-5}\cmidrule(l){6-8}
Task & Summary & Layers & \multicolumn{2}{c}{$\Delta_\ell$: median [Q1, Q3]} & Layers & \multicolumn{2}{c}{$\Delta_\ell$: median [Q1, Q3]} \\
\midrule
\multirow{3}{*}{\unigramlp} & $\tau_{\text{1st}}$ & 8/0/6 & $-1.72$ & [$-1.72$, $+0.35$] & 20/0/4 & $-1.04$ & [$-1.21$, $-0.86$] \\
 & $\tau_{\max}$ & 1/0/13 & $+5.19$ & [$+3.97$, $+5.30$] & 5/0/19 & $+5.17$ & [$+1.04$, $+5.87$] \\
 & $\tau_{50}$ & 0/1/13 & $+2.49$ & [$+1.71$, $+3.69$] & 1/0/23 & $+4.74$ & [$+1.57$, $+5.06$] \\
\midrule
\multirow{3}{*}{\bigramlp} & $\tau_{\text{1st}}$ & 16/0/0 & $-3.79$ & [$-4.66$, $-2.43$] & 24/0/0 & $-5.17$ & [$-5.35$, $-3.12$] \\
 & $\tau_{\max}$ & 14/0/2 & $-3.44$ & [$-3.96$, $-2.43$] & 19/0/5 & $-2.40$ & [$-3.12$, $-0.35$] \\
 & $\tau_{50}$ & 10/2/4 & $-1.57$ & [$-2.43$, $+0.37$] & 13/3/8 & $-0.79$ & [$-1.86$, $+0.45$] \\
\midrule
\multirow{3}{*}{\bigrampmi} & $\tau_{\text{1st}}$ & 15/0/1 & $-2.40$ & [$-2.40$, $-0.35$] & 20/0/4 & $-1.04$ & [$-2.40$, $-0.35$] \\
 & $\tau_{\max}$ & 14/0/2 & $-2.40$ & [$-2.40$, $-0.35$] & 19/0/5 & $-0.69$ & [$-2.40$, $-0.35$] \\
 & $\tau_{50}$ & 12/4/0 & $-0.82$ & [$-1.26$, $-0.43$] & 18/6/0 & $-0.66$ & [$-2.14$, $-0.39$] \\
\midrule
\multirow{3}{*}{\trigramlp} & $\tau_{\text{1st}}$ & 15/0/0 & $-1.04$ & [$-2.43$, $-1.04$] & 23/0/1 & $-1.72$ & [$-3.29$, $-1.72$] \\
 & $\tau_{\max}$ & 10/0/5 & $-0.35$ & [$-1.04$, $+2.40$] & 17/0/7 & $-1.37$ & [$-1.72$, $+2.58$] \\
 & $\tau_{50}$ & 6/1/8 & $+1.24$ & [$-0.44$, $+2.09$] & 10/1/13 & $+1.61$ & [$-0.88$, $+2.23$] \\
\midrule
\multirow{3}{*}{\trigrampmi} & $\tau_{\text{1st}}$ & 10/0/6 & $-0.35$ & [$-1.04$, $+0.51$] & 20/0/4 & $-1.04$ & [$-1.72$, $-0.35$] \\
 & $\tau_{\max}$ & 5/0/11 & $+0.68$ & [$-0.35$, $+3.10$] & 11/0/13 & $+0.35$ & [$-1.72$, $+1.04$] \\
 & $\tau_{50}$ & 3/1/12 & $+1.67$ & [$+0.48$, $+2.92$] & 10/3/11 & $+0.10$ & [$-1.17$, $+1.31$] \\
\midrule
\multirow{3}{*}{\pos} & $\tau_{\text{1st}}$ & 3/0/13 & $+1.04$ & [$+0.35$, $+2.40$] & 6/7/11 & $+0.33$ & [$-0.34$, $+1.03$] \\
 & $\tau_{\max}$ & 3/0/13 & $+1.04$ & [$+0.35$, $+2.40$] & 6/7/11 & $+0.33$ & [$-0.34$, $+1.03$] \\
 & $\tau_{50}$ & 1/2/13 & $+1.85$ & [$+0.53$, $+2.66$] & 1/10/13 & $+0.55$ & [$+0.10$, $+0.99$] \\
\midrule
\multirow{3}{*}{\deprel} & $\tau_{\text{1st}}$ & 5/5/6 & $-0.33$ & [$-0.35$, $+1.03$] & 5/19/0 & $-0.33$ & [$-0.33$, $+0.33$] \\
 & $\tau_{\max}$ & 5/3/8 & $+0.35$ & [$-0.35$, $+1.03$] & 5/19/0 & $-0.33$ & [$-0.33$, $+0.33$] \\
 & $\tau_{50}$ & 0/5/11 & $+0.85$ & [$+0.12$, $+1.22$] & 5/4/15 & $+0.73$ & [$-0.19$, $+1.05$] \\
\midrule
\multirow{3}{*}{\deplen} & $\tau_{\text{1st}}$ & 14/2/0 & $-1.72$ & [$-1.72$, $-1.04$] & 17/7/0 & $-2.41$ & [$-3.11$, $-0.33$] \\
 & $\tau_{\max}$ & 5/6/5 & $+0.33$ & [$-1.03$, $+1.72$] & 5/14/5 & $-0.33$ & [$-0.33$, $+0.33$] \\
 & $\tau_{50}$ & 3/4/9 & $+0.51$ & [$-0.03$, $+1.33$] & 7/5/12 & $+0.31$ & [$-0.44$, $+0.62$] \\
\midrule
\multirow{3}{*}{\depdepth} & $\tau_{\text{1st}}$ & 9/4/3 & $-0.35$ & [$-1.37$, $+0.33$] & 21/1/2 & $-3.11$ & [$-3.11$, $-1.73$] \\
 & $\tau_{\max}$ & 1/4/11 & $+2.07$ & [$+0.33$, $+2.41$] & 6/7/11 & $+0.33$ & [$-0.01$, $+3.97$] \\
 & $\tau_{50}$ & 0/4/12 & $+1.99$ & [$+0.35$, $+2.21$] & 7/6/11 & $+0.21$ & [$-0.37$, $+2.69$] \\
\midrule
\multirow{3}{*}{\ner} & $\tau_{\text{1st}}$ & 9/3/4 & $-0.35$ & [$-1.72$, $-0.16$] & 8/15/1 & $-0.33$ & [$-1.72$, $+0.33$] \\
 & $\tau_{\max}$ & 6/6/4 & $-0.33$ & [$-0.35$, $-0.16$] & 1/20/3 & $+0.33$ & [$-0.33$, $+0.33$] \\
 & $\tau_{50}$ & 2/9/5 & $+0.09$ & [$0.00$, $+0.41$] & 1/3/20 & $+1.28$ & [$+0.73$, $+1.62$] \\
\midrule
\multirow{3}{*}{\ssnoun} & $\tau_{\text{1st}}$ & 15/1/0 & $-1.03$ & [$-2.40$, $-0.35$] & 14/10/0 & $-1.02$ & [$-1.03$, $-0.33$] \\
 & $\tau_{\max}$ & 14/1/1 & $-0.69$ & [$-1.03$, $-0.35$] & 12/11/1 & $-0.34$ & [$-1.03$, $-0.33$] \\
 & $\tau_{50}$ & 6/6/4 & $-0.24$ & [$-1.02$, $+0.29$] & 0/13/11 & $+0.32$ & [$+0.11$, $+0.63$] \\
\midrule
\multirow{3}{*}{\ssverb} & $\tau_{\text{1st}}$ & 11/4/1 & $-1.72$ & [$-5.58$, $-0.33$] & 23/0/1 & $-5.38$ & [$-6.27$, $-2.40$] \\
 & $\tau_{\max}$ & 9/5/2 & $-1.72$ & [$-4.76$, $-0.33$] & 23/0/1 & $-5.38$ & [$-6.27$, $-1.54$] \\
 & $\tau_{50}$ & 10/5/1 & $-0.70$ & [$-5.19$, $-0.26$] & 20/2/2 & $-4.10$ & [$-5.56$, $-1.34$] \\
\midrule
\multirow{3}{*}{\emph{All tasks}} & $\tau_{\text{1st}}$ & 130/19/40 & $-1.04$ & [$-2.40$, $-0.33$] & 201/59/28 & $-1.04$ & [$-3.11$, $-0.33$] \\
 & $\tau_{\max}$ & 87/25/77 & $-0.33$ & [$-1.04$, $+1.04$] & 129/78/81 & $-0.33$ & [$-1.71$, $+0.35$] \\
 & $\tau_{50}$ & 53/44/92 & $+0.27$ & [$-0.46$, $+1.69$] & 93/56/139 & $+0.31$ & [$-0.65$, $+1.25$] \\
\bottomrule
\end{tabular}
}
\label{tab:event-summaries}
\end{table}

\subsection{Number of qualifying peaks per layer}
\label{app:subsec:multipeak}

As discussed in \cref{subsec:multipeak}, under the rule of qualifying peak in \cref{app:peak-rule}, \cetext can have more than one peak, and the numbers of peaks on different layers and tasks vary widely.
Specifically, across all tasks, 44.8\% of layer trajectories in Pythia-410M have more than one peak, and 45.8\% in Pythia-1B.
\Cref{tab:multipeak} shows the statistics of the number of peaks across tasks for Pythia-410M and -1B.
The multi-peak phenomenon is more common in n-gram tasks: in Pythia-410M, \bigramlp has multiple peaks in 79.2\% of layers and \unigramlp in 75.0\%; in Pythia-1B, \bigramlp has 75.0\%, followed by \trigramlp and \trigrampmi.
This is less common in classification tasks (Pythia-410M and -1B: 35.7\% and 38.4\% of layers, vs.\ 57.5\% and 56.3\% for n-gram tasks), with large differences between tasks.
For example, for \pos, all layers in both models have only a single peak, and for \deprel the multi-peak share is 8.3\% (410M) and 12.5\% (1B).
However, some classification tasks are predominantly multi-peak, such as \depdepth at 75.0\% in Pythia-410M and 62.5\% in Pythia-1B.

\begin{table}[t]\centering\small
    \caption{Distribution of qualifying interior \cetext peaks per layer under the gate of \Cref{app:peak-rule}.
    Entries are percentages of all layers, including layers without a qualifying peak; ``All tasks'' pools the layers of the 12 tasks.
    Rows may not sum to 100 owing to rounding.}
    \begin{tabular}{lrrr}
\toprule
Task & 0 peaks & 1 peak & $\geq 2$ peaks \\
\midrule
\textbf{Pythia-410M} (24 layers) & & & \\
\unigramlp & 0.0 & 25.0 & 75.0 \\
\bigramlp & 0.0 & 20.8 & 79.2 \\
\bigrampmi & 0.0 & 79.2 & 20.8 \\
\trigramlp & 0.0 & 45.8 & 54.2 \\
\trigrampmi & 0.0 & 41.7 & 58.3 \\
\pos & 0.0 & 100.0 & 0.0 \\
\deplen & 0.0 & 50.0 & 50.0 \\
\deprel & 0.0 & 91.7 & 8.3 \\
\depdepth & 0.0 & 25.0 & 75.0 \\
\ner & 0.0 & 58.3 & 41.7 \\
\ssnoun & 0.0 & 79.2 & 20.8 \\
\ssverb & 0.0 & 45.8 & 54.2 \\
All tasks & 0.0 & 55.2 & 44.8 \\
\midrule
\textbf{Pythia-1B} (16 layers) & & & \\
\unigramlp & 12.5 & 43.8 & 43.8 \\
\bigramlp & 0.0 & 25.0 & 75.0 \\
\bigrampmi & 0.0 & 62.5 & 37.5 \\
\trigramlp & 6.2 & 31.2 & 62.5 \\
\trigrampmi & 0.0 & 37.5 & 62.5 \\
\pos & 0.0 & 100.0 & 0.0 \\
\deplen & 0.0 & 43.8 & 56.2 \\
\deprel & 0.0 & 87.5 & 12.5 \\
\depdepth & 0.0 & 37.5 & 62.5 \\
\ner & 0.0 & 50.0 & 50.0 \\
\ssnoun & 0.0 & 68.8 & 31.2 \\
\ssverb & 0.0 & 43.8 & 56.2 \\
All tasks & 1.6 & 52.6 & 45.8 \\
\bottomrule
\end{tabular}

\label{tab:multipeak}
\end{table}

\subsection{Rules for qualifying energy events}
\label{app:peak-rule}

After min-max-normalizing each within-layer \([\ce_{\ell,t}]_+\) curve to \([0,1]\), let \(p^{\mathrm{high}}\) and \(p^{\mathrm{low}}\) be the rises from the higher and lower adjacent valleys at a candidate peak.
Under our default criterion, a candidate qualifies when \(p^{\mathrm{high}}\geq 0.05\) (rejecting grid-noise peaks) and either \(p^{\mathrm{high}}\geq 0.25\) (symmetric peak) or \(p^{\mathrm{low}}\geq 0.75\) (one-sided cliff); among qualifying candidates, the earliest is selected.
We refer to this criterion as \(0.05/0.25/0.75\).
Only interior peaks are eligible; endpoints are excluded.
If no candidate passes the gate, the energy event is undefined.
A layer is valid for paired timing analysis only when both
the energy event and the decodability transition are defined.
Layers with undefined events are excluded from timing summaries, but their trajectories remain in the full-layer plots (\cref{fig:ce-heatmap-410m} and \cref{fig:ce-heatmap-1b}).

\section{Probe reliability}
\label{app:probe-reliability}

We conducted controlled experiments on the POS task using the Pythia-410M model to examine the sensitivity of the \cetext pattern to the reuse of fitted probe weights, the number of probe directions $k$, and probe quality (as influenced by training data size).
All analyses utilized the same 19 evaluation checkpoints as the main experiment and adopted the same \cetext definitions and event filtering rules (see \Cref{app:peak-rule}).

\subsection{Independently fitted probes}
\label{app:independent-probes}

In our analysis, the fitted probe defines both the concept subspace and its decodability curve.
To assess \cetext's sensitivity to this reuse, we split the official training set into two sentence-disjoint subsets (A with 88,364 words and B with 88,474 words) and trained both real and control POS probes (3 seeds each) on each subset at all layers of Pythia-410M.
Both subsets employed the same training protocol, with input hidden states mean-centered using each subset's respective training mean, and a shared official held-out test set was used to evaluate decodability.
We constructed subspaces $U_A$ and $U_B$ using the respective probe weights and conducted two cross-fit tests: comparing the \cetext derived from $U_A$ against the decodability of probe B, and conversely, comparing the \cetext derived from $U_B$ against the decodability of probe A.

\begin{figure}[!tbp]
  \centering
  \includegraphics[width=\linewidth]{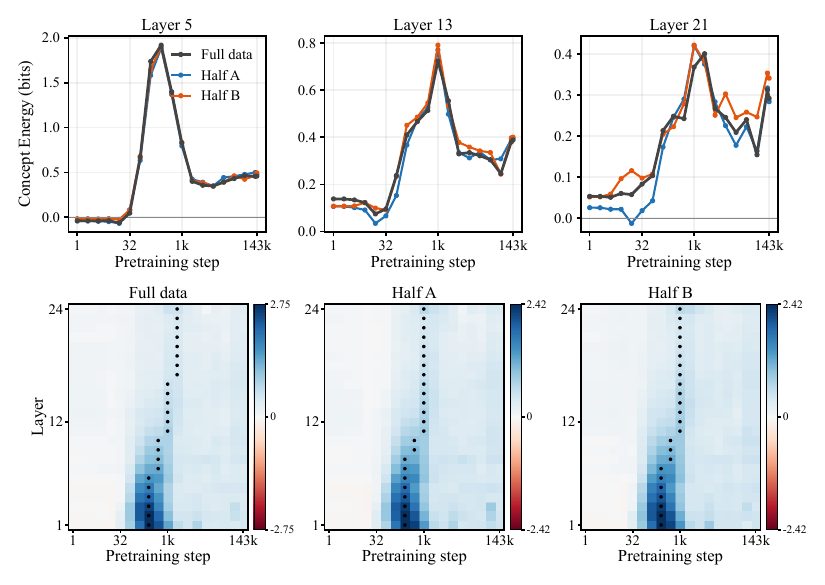}
  \caption{\cetext of \pos probes in Pythia-410M fitted on the full training set, on half A, and on half B.
  Top: layers 5, 13, and 21, the layers of \Cref{app:probe-data-sensitivity}.
  Bottom: all 24 layers (rows, numbered from 1) and 19 checkpoints (columns, equally spaced); blue: \(\ce>0\); red: \(\ce<0\); darker cells have larger \(|\ce|\), and each panel has its own symmetric scale.
  Dots mark the energy event $t^E_\ell$ of each layer.
  Each \cetext uses the real and control probes fitted on the same data, averaged over 3 real and 3 control seeds.}
  \label{fig:crossfit}
\end{figure}

As shown in \Cref{tab:crossfit}, in both the A$\to$B and B$\to$A directions, the energy event and the decodability transition event coincide (with event lags smaller than $\ln 2/2$ in both cases).
The top row of \cref{fig:crossfit} shows how \cetext changes in this cross-fit experiment at layers 5, 13, and 21: the \cetext curves obtained from probes trained separately on the two halves are highly consistent with the full-data \cetext curve.
In addition, these independently fitted subspaces yielded valid energy events across all 24 layers of Pythia-410M.
The median Spearman correlation between the two probes' \cetext trajectories is 0.93, with quartiles of $[0.88, 0.96]$, and 23 of 24 layers yielding the same event steps.
The only discrepancy occurred at layer 8: the subspace derived from A selected step 256, whereas the subspace derived from B selected step 512.
This consistency supports the reproducibility of energy event selection during independent probe fitting under this setup.

\begin{table}[t]\centering\small
    \caption{Cross-fit event timing for \pos in Pythia-410M.
    Each row pairs the energy event from one set of probes with the decodability event from another: half A with half B (A$\to$B), half B with half A (B$\to$A), and, for reference, the probes fitted on the full training set (main text).
    $t^D$ and $t^E$ are aggregated over the 24 layers by \cref{eq:aggregated-event-time}; in all three rows $|\ln(t^E/t^D)|<\ln 2/2$, so the two events coincide.}
    \begin{tabular}{lrrl}
\toprule
Pairing & $t^D$ & $t^E$ & $\ln(t^E/t^D)$ \\
\midrule
CE from half A, selectivity from half B & 370 & 421 & $+0.13$ (coincide) \\
CE from half B, selectivity from half A & 382 & 438 & $+0.14$ (coincide) \\
Full data (main text) & 493 & 479 & $-0.03$ (coincide) \\
\bottomrule
\end{tabular}

\label{tab:crossfit}
\end{table}

\subsection{Sensitivity to the number of probe directions}
\label{app:k-sensitivity}

Different tasks have different numbers of probe directions $k$ (\cref{app:probe-details}).
To examine whether \(\ce\) depends on $k$, we directly reuse the \pos probes and corresponding control probes from \Cref{app:independent-probes}, replacing each subspace with the top $k$ right singular vectors of the row-centered weight matrix for $k=1,4,8,12$ ($k=15$ is the full probe).

\begin{figure}[!tbp]
  \centering
  \includegraphics[width=\linewidth]{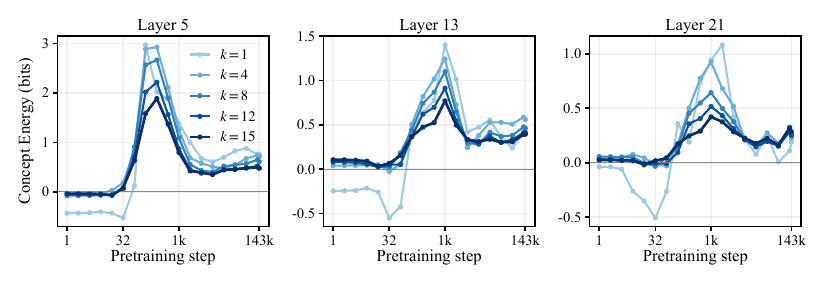}
  \caption{\cetext of the half-A \pos probes in Pythia-410M restricted to their top $k$ directions, at layers 5, 13, and 21.
  $k=15$ is the full probe (half A in \Cref{fig:crossfit}); real and control probes use the same $k$.
  Each curve is the mean over 3 real and 3 control seeds.}
  \label{fig:k-sweep}
\end{figure}

As shown in \Cref{fig:k-sweep}, smaller $k$ generally yields larger \(\ce\) values (especially around the peak interval at steps 256--2k): since \oetext compares the subspace with a random subspace of the same rank, it averages over directions, and the leading directions of the real probe carry more AGOP energy than the later ones, whereas the directions of the control probe differ little;
meanwhile, the shape of the \cetext trajectory and the location of the energy event remain largely unchanged.

When $k\ge 4$, the median Spearman $\rho$ between each layer's trajectory and the $k=15$ trajectory is at least 0.93. Across different $k$, $t^E$ within the same layer changes very little (remaining mostly unchanged for $k\ge 4$, and shifting earlier or later by mostly one log step at $k=1$), and the energy event and decodability event still coincide under both cross-fit pairings.
At $k=1$, the trajectories are noticeably noisier with a lower signal-to-noise ratio: in the last two columns of \Cref{tab:k-sweep}, since $\sigma_0\propto 1/\sqrt{k}$, the noise at $k=1$ is about $3.9\times$ that at $k=15$, and the ratio of \(\ce\) to $\sigma_0$ drops from about 8 to about 3.

\begin{table}[t]\centering\small
    \caption{Sensitivity of \cetext to the number of probe directions $k$ for \pos in Pythia-410M ($k=15$ is the full probe).
    $\rho$: Spearman correlation between the \cetext trajectories at $k$ and at $k=15$ over all 48 trajectories (2 halves $\times$ 24 layers); ``$t^E$ unchanged'': share of trajectories with defined events at both ranks (47 for $k=4,8$ and 48 otherwise) whose energy event is at the same step as at $k=15$.
    Cross-fit lag: $\ln(t^E/t^D)$ of the two cross-fit pairings of \Cref{tab:crossfit}.
    $\sigma_0$: null standard deviation of the energy of $k$ random directions (\Cref{app:sigma0}); $\mathrm{CE}/\sigma_0$ over cells from step 128 on.
    Entries are median [IQR].}
    \resizebox{\linewidth}{!}{\begin{tabular}{rrrrrr}
\toprule
$k$ & $\rho$ vs.\ $k=15$ & $t^E$ unchanged & Cross-fit (A$\to$B / B$\to$A) & $\sigma_0$ (bits) & $\mathrm{CE}/\sigma_0$ \\
\midrule
1 & 0.85 [0.82, 0.89] & 42\% & $+0.20$ / $-0.08$ & 0.21 [0.17, 0.26] & 3.04 [1.88, 4.66] \\
4 & 0.93 [0.89, 0.95] & 81\% & $+0.25$ / $+0.16$ & 0.11 [0.09, 0.13] & 6.37 [4.50, 8.38] \\
8 & 0.96 [0.93, 0.98] & 98\% & $+0.17$ / $+0.12$ & 0.08 [0.06, 0.09] & 7.33 [5.35, 9.79] \\
12 & 0.97 [0.95, 0.99] & 94\% & $+0.12$ / $+0.15$ & 0.06 [0.05, 0.07] & 8.20 [5.59, 10.58] \\
15 & 1.00 [1.00, 1.00] & 100\% & $+0.13$ / $+0.14$ & 0.05 [0.04, 0.07] & 8.43 [5.54, 11.22] \\
\bottomrule
\end{tabular}
}
\label{tab:k-sweep}
\end{table}

\subsection{Sensitivity to probe-training data}
\label{app:probe-data-sensitivity}

To evaluate the sensitivity of \cetext to probe quality, we varied the amount of labeled POS data used to fit the probes while keeping other conditions unchanged.
We assessed layers 5, 13, and 21 of the Pythia-410M model to cover shallow, middle, and deep layers.

\begin{figure}[htbp]
    \centering
    \includegraphics[width=\linewidth]{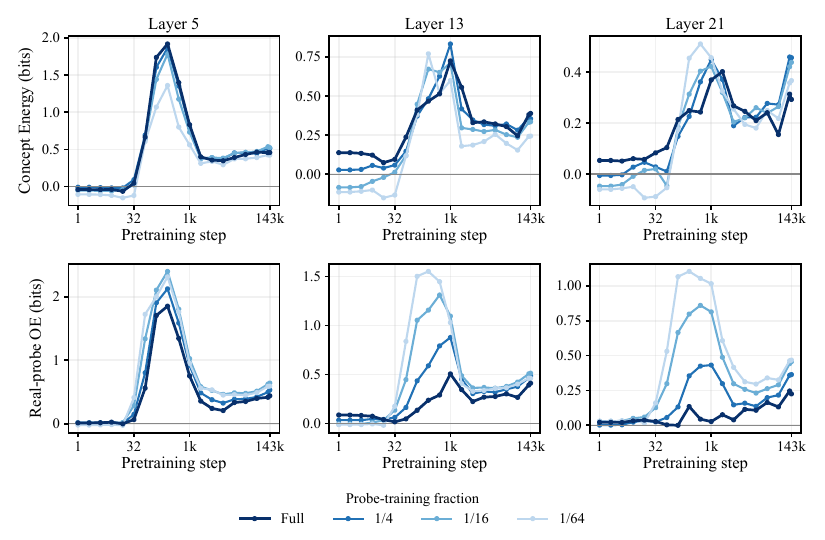}
    \caption{\cetext (top) and the \oetext of the real probes alone (bottom) for \pos probes in Pythia-410M fitted on different fractions of the training data, at layers 5, 13, and 21 (columns).
    Each reduced fraction is averaged over 3 random subsamples of the training set; the full-data curve is the main-text time course.}
    \label{fig:training_fraction_sensitivity}
\end{figure}

Reduced training sets were constructed using three subsample seeds (42, 43, and 44).
For each seed, we sampled approximately 1/4, 1/16, and 1/64 of the data from each POS class to preserve their original proportions; smaller training sets were always subsets of the larger ones. 
For each reduced-data condition, at each selected layer and checkpoint we train one real probe and one control probe on each of the three data-subsampling seeds (with training randomness controlled by seeds 1001 and 4042, respectively), average the resulting \oetext across the subsamples, take their difference to obtain \cetext, and finally compare against the corresponding full-data baseline (averaged over 3 real probes and 3 control probes as in \cref{app:probe-details}) at the same layer and the same checkpoint.
Note that when mean-centering the inputs before training the reduced-fraction probes, each subset uses its own training-set mean, whereas the control task's label mapping is determined once on the full training set and applied directly to the different training-fraction subsets.
The results (\Cref{tab:training_fraction_sensitivity} and \Cref{fig:training_fraction_sensitivity}) show that the \(\ce\) magnitude is more robust to the training fraction than the \(\oem\) magnitude; the latter also exhibits lower overall Spearman correlations across different training fractions than the former.

\begin{table}[t]\centering\small
\caption{Sensitivity of \cetext to the amount of probe-training data for \pos in Pythia-410M.
Each reduced fraction uses 3 random subsamples of the training set, and the \oetext of the real and of the control probes is averaged over the subsamples before taking their difference; the full row is the main-text time course.
Held-out acc.: median over the 19 checkpoints of the test accuracy of the real probes.
Peak / full: the maximum of the trajectory over checkpoints divided by that with the full data; $\rho$: Spearman correlation between the trajectory and that with the full data.
The \oetext columns use the real probes alone, without subtracting the control.
Entries are median [min, max] over layers 5, 13, and 21.}
\resizebox{\linewidth}{!}{\begin{tabular}{lrrrrr}
\toprule
Train fraction & Held-out acc. & Peak CE / full & $\rho$ (CE) & Peak OE / full & $\rho$ (OE) \\
\midrule
Full & 0.82 [0.82, 0.82] & 1.00 [1.00, 1.00] & 1.00 [1.00, 1.00] & 1.00 [1.00, 1.00] & 1.00 [1.00, 1.00] \\
$1/4$ & 0.80 [0.80, 0.81] & 1.14 [0.97, 1.15] & 0.93 [0.87, 0.96] & 1.73 [1.15, 1.76] & 0.79 [0.64, 1.00] \\
$1/16$ & 0.77 [0.77, 0.77] & 0.98 [0.93, 1.09] & 0.91 [0.86, 0.95] & 2.59 [1.30, 3.50] & 0.63 [0.29, 0.96] \\
$1/64$ & 0.72 [0.71, 0.73] & 1.06 [0.71, 1.27] & 0.88 [0.86, 0.96] & 3.07 [1.25, 4.49] & 0.55 [0.21, 0.98] \\
\bottomrule
\end{tabular}
}
  \label{tab:training_fraction_sensitivity}
\end{table}

\section{Selectivity and raw probe score}
\label{app:selectivity}

\begin{figure}[!tbp]
  \centering
  \includegraphics[width=\linewidth]{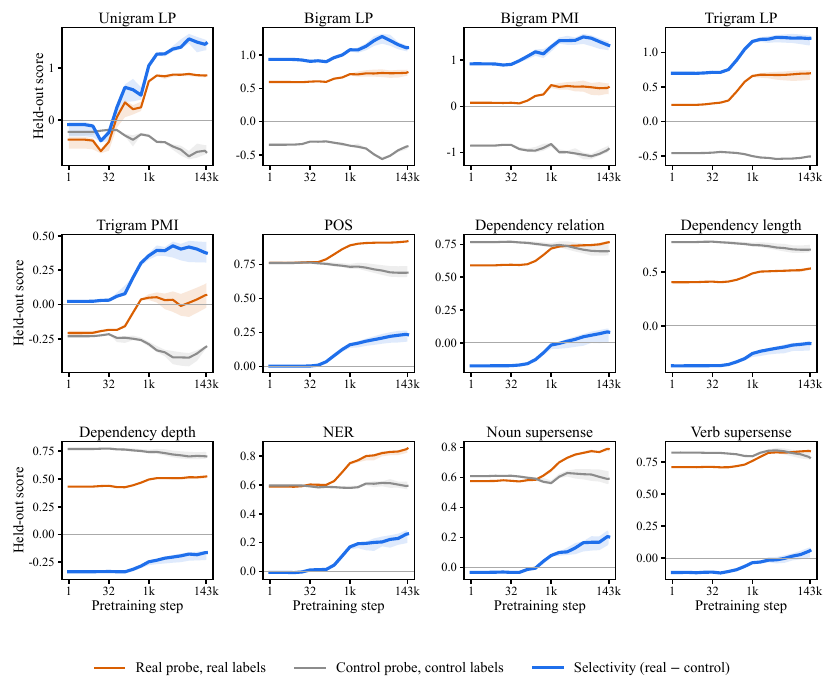}
  \caption{Held-out scores over pretraining for the 12 tasks in Pythia-1B: the real probe on the real labels (orange), the control probe on its own control labels (gray), and selectivity, their difference (blue).
  Scores are accuracy for the classification tasks and \(R^2\) for the regression (n-gram) tasks; the first five panels are the regression tasks.
  Lines and bands: median and 25th--75th percentiles across the 16 layers, each layer averaged over probe seeds.
  Y-axes differ across panels.}
  \label{fig:selectivity-scores-1b}
\end{figure}

\begin{figure}[!tbp]
  \centering
  \includegraphics[width=\linewidth]{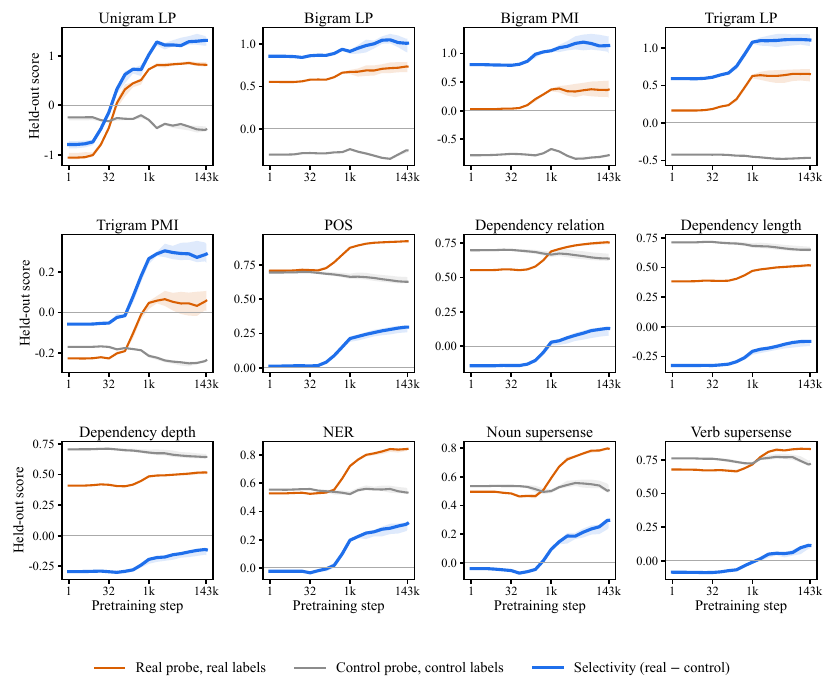}
  \caption{As \Cref{fig:selectivity-scores-1b}, for Pythia-410M: held-out scores of the real probe, the control probe and selectivity, with the median and 25th--75th percentiles across the 24 layers.}
  \label{fig:selectivity-scores-410m}
\end{figure}

\Cref{sec:theoretical-method} operationalizes decodability as selectivity (\Cref{subsec:concept-subspaces}).
In this section, we show the held-out scores of the real probe and the control probe on their respective labels.
Classification tasks use accuracy, and n-gram tasks use held-out $R^2$ (\Cref{fig:selectivity-scores-1b,fig:selectivity-scores-410m}).
Compared with the real-task probe scores, the control-probe scores change very little throughout training.
For many tasks, the shape of the selectivity curve is nearly a vertical shift of the real-score curve (e.g., \bigrampmi, \trigramlp, \pos, and \ner), so selectivity is largely determined by the real-probe score.

For all n-gram tasks, the control $R^2$ is negative throughout training because the training and test sets are split to have disjoint n-gram types (\Cref{app:probe-details}).
Consequently, the control probe cannot predict unseen types.
As a result, the selectivity of the n-gram tasks exceeds the score of the real probe itself.

In both models, the selectivity of \deplen and \depdepth is negative at all checkpoints.
This is because the control labels are determined by word type, and word identity can be easily read out linearly from the residual stream.
By contrast, dependency arc length and depth depend more on syntactic context than on word identity, i.e., on these two tasks, memorizing by word type is easier to decode than the true labels; these two factors together lead to negative selectivity.
Even for these two tasks, however, selectivity rises steadily during training.
Because identifying the decodability event depends only on the interval with steepest slope (\cref{eq:transition-event}), a decodability event can still be found even when selectivity is negative overall.
In spite of that, we advocate interpreting \cetext and its timing relative to the decodability event conservatively when selectivity is negative overall.

If we use the real-probe score instead of selectivity to compute $t^D$, the timing-order counts are \texttt{10/2/0} (Pythia-410M) and \texttt{10/1/1} (Pythia-1B), compared with \texttt{11/1/0} and \texttt{9/2/1} when using selectivity (\cref{subsec:timing-diagonal}).
Specifically, under this setting, \deprel in Pythia-1B changes from coinciding to energy-earlier, and \unigramlp in Pythia-410M changes from energy-earlier to coinciding.

\section{Behavioral experiments}
\label{app:behavior}

\begin{figure}[!htbp]
  \centering
  \includegraphics[width=\linewidth]{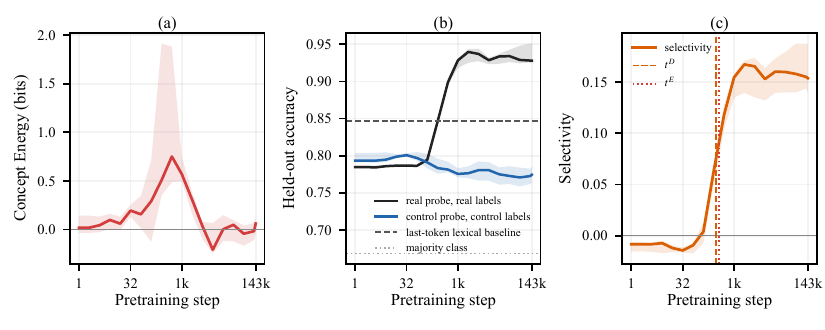}
  \caption{
    Subject--verb agreement in Pythia-410M over pretraining: \cetext (left), held-out accuracy of the real and control probes on their respective labels (middle), and selectivity (right).
    Lines and bands show the median and 25th--75th percentiles across the 24 layers, with each layer averaged over three probe seeds.
    Middle: the dashed line is the last-token lexical baseline (0.847) and the dotted line the majority class (0.669).
    Right: vertical lines mark $t^D$ and $t^E$ (\cref{eq:aggregated-event-time}).
  }
  \label{fig:app-sva}
\end{figure}

\paragraph{Subject--verb agreement.}
For the subject--verb agreement (SVA) probe, we start from the corpus of \citet{linzen-etal-2016-assessing} (about 1.58M sentences), where singular and plural verb labels come from the original \texttt{VBZ}/\texttt{VBP} POS tags.
Following the subset filtering of \citet{goldberg2019assessing}, we first keep sentences with 1--4 attractors (71{,}704 sentences) and then remove sentences whose target verb is \emph{is} or \emph{are}, leaving 29{,}985 sentences (33.1\% singular; identical to the file in \citeauthor{goldberg2019assessing}'s repository).
We group these sentences by prefix and split them into a training set (\textasciitilde 80\%, 23{,}987 sentences) and a held-out test set (5{,}998 sentences) so that the same prefix never appears in both splits; decodability is measured by test selectivity (\Cref{fig:app-sva}).
In the control task, the label is tied to the final token of the prefix, ignoring the true grammatical number in the context.
This setup defines a natural lexical baseline on the test set: looking only at the final token of the prefix, we predict its majority label if that token appeared in the training set, and predict the overall majority class otherwise.
This lexical baseline achieves an accuracy of 0.847, compared with 0.669 for always guessing the majority class.
To measure model behavior, we use the macro-average accuracy across the six paradigms in the subject--verb agreement category of BLiMP, and we compute the behavioral transition event $t^B$ as the point of steepest increase on this curve (\cref{eq:transition-event}).

\begin{figure}[!htbp]
  \centering
  \includegraphics[width=\linewidth]{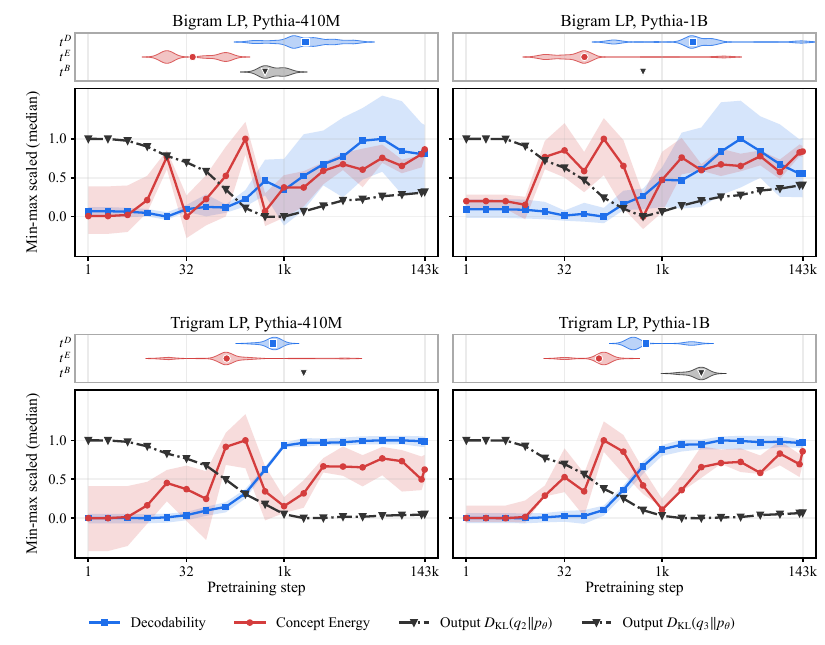}
  \caption{
    Decodability (selectivity; blue), \cetext (red), and output divergence $D_{\mathrm{KL}}(q_n\|p_\theta)$ (black) for bigrams (top) and trigrams (bottom) in Pythia-410M (left) and Pythia-1B (right), drawn as in \Cref{fig:sva-decode-energy-behavior}(b).
    Lines are medians over layers, min--max scaled; bands span the nine pairs of real- and control-probe seeds; the divergence has no band.
    The strip above each panel marks $t^D$, $t^E$, and $t^B$ as in \Cref{fig:sva-decode-energy-behavior}; $t^B$ has a violin only for Pythia-410M bigrams and Pythia-1B trigrams, the two cases in which no single checkpoint accounts for at least 95\% of the 1{,}000 bootstrap minima.
  }
  \label{fig:app-bergen}
\end{figure}

\paragraph{N-gram output behavior.}
For the n-gram output behavioral analysis (\Cref{fig:app-bergen}), we sample 1{,}000 documents from the 29{,}436 documents in the WikiText-103 training split using a fixed random seed.
From each document, we take one random window of up to 128 tokens, giving about 125{,}000 next-token prediction positions in total.
At each position, we compare the reference n-gram distribution $q_n$ (estimated from WikiText-103) with the model's next-token distribution $p_\theta$.
Following \citet{chang-bergen-2022-word}, we compute the divergence $D_{\mathrm{KL}}(q_n\|p_\theta)$ in bits (log base 2) at each position and average it over all valid positions for each checkpoint.
We do not back off to lower-order n-grams when a context is absent from the reference n-gram vocabulary.
Instead, we exclude positions whose context does not appear in the reference table, which retains 100\% of positions for unigrams, 99.0\% for bigrams, and 97.9\% for trigrams.
The excluded positions depend only on the reference vocabulary and the evaluation text, not on the model, so all 19 checkpoints and both models are evaluated on the exact same set of positions for each n-gram order.
To evaluate the stability of $t^B$, we resample the 1{,}000 documents with replacement 1{,}000 times and recompute the checkpoint of minimum KL divergence; the distribution of these minima is shown as the $t^B$ violins in \Cref{fig:sva-decode-energy-behavior,fig:app-bergen}.
Across both models and all three n-gram orders, the timing-order count between the energy event $t^E$ and the checkpoint of minimum KL divergence $t^B$ is \texttt{6/0/0} (\texttt{earlier/same/later}); i.e., the energy event occurs earlier in all six cases.

\section{Compute resources, existing assets, and licenses}
\label{app:compute-licenses}

\paragraph{Compute resources.}
All experiments were run on single-GPU instances on Jetstream2 with a persistent 1 TB attached volume for model caches, datasets, AGOP caches, and probe outputs.
AGOP precomputation was run in fp16 with batch size 4 on an NVIDIA A100 20 GiB partition (\texttt{GRID A100X-20C}) with 16 AMD EPYC-Milan vCPUs and about 58 GiB of host memory, using about 17--18 GiB of peak GPU memory on Pythia-1B.
Computing the shared AGOP cache across all 19 checkpoints took about 2.6 GPU-hours for all 16 layers of Pythia-1B together with the final layer of Pythia-410M, and about 6 GPU-hours for the remaining 23 layers of Pythia-410M, totaling roughly 9 GPU-hours.

Probe training, control tasks, and behavioral evaluations reuse the shared AGOP cache and were run on full NVIDIA A100-SXM4-40GB GPUs with 32 vCPUs and about 115 GiB of host memory.
Each unit trained real and control probes (3 seeds each) at one checkpoint and a group of layers, using at most 3.9 GiB of peak GPU memory and typically 22--33 GiB of host memory (up to 59 GiB for Pythia-1B trigram units, which load 500{,}000 token positions across five layers at once).
Because 5--8 probe jobs typically shared a single GPU concurrently, we report conservative wall-clock GPU-hours:
7-task classification sweeps took 19.6 GPU-hours on Pythia-410M and 40.7 GPU-hours on Pythia-1B;
5-task n-gram sweeps across both models took 45.0 GPU-hours;
uncentered-input ablations on Pythia-410M (\Cref{app:centring}) took 28.1 GPU-hours;
cross-fit, $k$-sensitivity, training-budget, and SVA/BLiMP experiments (\Cref{app:probe-reliability,app:behavior}) took 10.8 GPU-hours;
and the n-gram KL divergence evaluations took under 1 GPU-hour (about 30 seconds per checkpoint).
In total, all experiments reported in the paper required approximately 155 GPU-hours (about 9 GPU-hours for AGOP and 145 GPU-hours for probe and behavioral runs).
Over the course of the project, including preliminary, superseded, and failed runs not reported in the paper, our Jetstream2 resources were in active use for roughly 730 hours, about 4--5$\times$ the compute of the reported experiments. This figure is estimated from hours with experiment-output writes and does not include earlier exploratory experiments on a shared institutional Kubernetes GPU cluster.

\paragraph{Existing assets and licenses.}
All models, datasets, and software packages used in this work are existing public research assets, and we cite their original publications in the main text.
The Pythia checkpoints~\citep{biderman2023pythia} are distributed by EleutherAI under the Apache License 2.0; the Pythia repository states that this license applies to the code, Pythia models, and other copyrightable artifacts in that repository.\footnote{\url{https://github.com/EleutherAI/pythia}}
The WikiText-103 corpus~\citep{merity2017pointer}, used for the lexical-statistic probes and AGOP corpus sampling, is distributed on HuggingFace as \texttt{Salesforce/wikitext}; its dataset card lists CC BY-SA 3.0 and GFDL.\footnote{\url{https://huggingface.co/datasets/Salesforce/wikitext}}
UD English EWT~\citep{silveira-etal-2014-gold,nivre-etal-2016-universal}, used for POS and dependency probes, is distributed by Universal Dependencies under CC BY-SA 4.0.\footnote{\url{https://universaldependencies.org/treebanks/en_ewt/index.html}}
UNER English EWT~\citep{mayhew-etal-2024-universal}, used for coarse NER, is distributed as part of \texttt{universalner/universal\_ner}; the HuggingFace dataset card lists CC BY-SA 4.0.\footnote{\url{https://huggingface.co/datasets/universalner/universal_ner}}
STREUSLE~\citep{schneider-smith-2015-corpus,schneider-etal-2018-comprehensive}, used for noun and verb supersense probes, is distributed under CC BY-SA 4.0 for its multiword-expression and supersense annotations; its repository notes that UD annotations are redistributed under the same license, while source sentences and PTB POS annotations from the English Web Treebank reviews section are redistributed with permission from Google and the Linguistic Data Consortium.\footnote{\url{https://github.com/nert-nlp/streusle}}
BLiMP~\citep{warstadt-etal-2020-blimp-benchmark}, used for the subject--verb agreement behavioral evaluation, is distributed under CC BY 4.0.\footnote{\url{https://github.com/alexwarstadt/blimp}}
The subject--verb agreement dataset~\citep{linzen-etal-2016-assessing,goldberg2019assessing} was extracted from Wikipedia (CC BY-SA) and released with the MIT-licensed \texttt{rnn\_agreement} repository and the Apache-2.0-licensed \texttt{bert-syntax} repository; no separate data license is stated.\footnote{\url{https://github.com/TalLinzen/rnn_agreement}; \url{https://github.com/yoavg/bert-syntax}}
Our implementation builds on PyTorch (BSD-3-Clause), HuggingFace Transformers (Apache License 2.0), and \texttt{nnsight} (MIT License).
We use these assets for noncommercial academic analysis and report aggregate probe and geometry statistics rather than redistributing raw corpus text in the paper.

\clearpage
\newpage
\section*{NeurIPS Paper Checklist}

\begin{enumerate}

\item {\bf Claims}
    \item[] Question: Do the main claims made in the abstract and introduction accurately reflect the paper's contributions and scope?
    \item[] Answer: \answerYes{}
    \item[] Justification: The abstract and \Cref{sec:intro} state four claims, supported in \Cref{sec:results,subsec:multipeak}: \cetext is positive for most tasks and layers in Pythia-410M and -1B (\Cref{subsec:ce-positive}); the earliest energy event precedes the steepest rise in decodability in 11 and 9 of the 12 probing tasks, with the exceptions listed (\Cref{subsec:timing-diagonal}); energy events precede the behavioral events for subject--verb agreement and for the n-gram output divergence (\Cref{subsec:behavior}); and \cetext trajectories often have more than one peak (\Cref{subsec:multipeak}). The claims are limited to these two models and these tasks (\Cref{subsec:limitations}).

\item {\bf Limitations}
    \item[] Question: Does the paper discuss the limitations of the work performed by the authors?
    \item[] Answer: \answerYes{}
    \item[] Justification: \Cref{subsec:limitations} states that \cetext measures loss-gradient geometry rather than causal reliance, that probe subspaces are operational definitions, that the analysis covers two sizes of one model family, and that the timing of energy events relative to behavior is tested only for tasks with a behavioral benchmark.

\item {\bf Theory assumptions and proofs}
    \item[] Question: For each theoretical result, does the paper provide the full set of assumptions and a complete (and correct) proof?
    \item[] Answer: \answerYes{}
    \item[] Justification: The paper has no theorem-style results. The mathematical statements the method relies on are derived with their assumptions in \Cref{app:math-derivations}: the chain-rule decomposition of next-token log-probability (\Cref{app:chain-rule-proof}), the random-subspace expectations used in \oetext (\Cref{app:haar-expectations}), and the noise scale of \oetext (\Cref{app:sigma0}); \Cref{app:centring} derives the gradient decomposition behind input and target centering.

\item {\bf Experimental result reproducibility}
    \item[] Question: Does the paper fully disclose all the information needed to reproduce the main experimental results of the paper to the extent that it affects the main claims and/or conclusions of the paper (regardless of whether the code and data are provided or not)?
    \item[] Answer: \answerYes{}
    \item[] Justification: \Cref{sec:experimental-methods} and \Cref{app:probe-details} specify the models, checkpoints, layers, tasks, dataset versions, data splits, control tasks, probe hyperparameters, AGOP sample size, sequence lengths, and sampled token positions, and \Cref{subsec:event-timing} defines the energy and decodability events; \Cref{app:centring,app:behavior} give the preprocessing and the behavioral evaluation protocols.

\item {\bf Open access to data and code}
    \item[] Question: Does the paper provide open access to the data and code, with sufficient instructions to faithfully reproduce the main experimental results, as described in supplemental material?
    \item[] Answer: \answerYes{}
    \item[] Justification: All models and datasets are public (\Cref{app:compute-licenses}). The code for the main experimental pipeline (AGOP computation, probe training, \cetext, and event extraction) is released at \url{https://github.com/YpLarryWang/concept-energy}, with instructions for obtaining the data and reproducing the main results.

\item {\bf Experimental setting/details}
    \item[] Question: Does the paper specify all the training and test details (e.g., data splits, hyperparameters, how they were chosen, type of optimizer) necessary to understand the results?
    \item[] Answer: \answerYes{}
    \item[] Justification: \Cref{sec:experimental-methods} and \Cref{app:probe-details} describe the frozen Pythia checkpoints, residual-stream layers, probe tasks, dataset filtering and splits, control tasks, optimizer, learning rate, weight decay, epochs, batch sizes, sequence lengths, and held-out evaluation metrics. The same probe hyperparameters are used for all tasks, layers, and checkpoints.

\item {\bf Experiment statistical significance}
    \item[] Question: Does the paper report error bars suitably and correctly defined or other appropriate information about the statistical significance of the experiments?
    \item[] Answer: \answerYes{}
    \item[] Justification: Probe scores and \oetext are averaged over three real-probe and three control-probe seeds. Curves show medians across layers with 25th--75th percentile bands, and in \Cref{fig:sva-decode-energy-behavior,fig:app-bergen} the bands span the nine pairs of real- and control-probe seeds. Layer-level event times are shown as distributions and summarized in \Cref{tab:event-summaries}. For the n-gram output divergence, we assess the stability of the minimum-divergence checkpoint with 1{,}000 document-bootstrap resamples (\Cref{app:behavior}). Sensitivity to independently fitted probes, the number of probe directions, the probe-training data, input centering, and the decodability measure is reported in \Cref{app:probe-reliability,app:centring,app:selectivity}. We do not perform formal hypothesis tests.

\item {\bf Experiments compute resources}
    \item[] Question: For each experiment, does the paper provide sufficient information on the computer resources (type of compute workers, memory, time of execution) needed to reproduce the experiments?
    \item[] Answer: \answerYes{}
    \item[] Justification: \Cref{app:compute-licenses} reports the GPU types, GPU and host memory, and wall-clock GPU-hours for each group of experiments (about 155 GPU-hours in total), as well as the compute used across the whole project, including runs not reported in the paper.

\item {\bf Code of ethics}
    \item[] Question: Does the research conducted in the paper conform, in every respect, with the NeurIPS Code of Ethics \url{https://neurips.cc/public/EthicsGuidelines}?
    \item[] Answer: \answerYes{}
    \item[] Justification: The work analyzes public model checkpoints and public annotated datasets. It collects no personal data, involves no human subjects, and releases no new model.

\item {\bf Broader impacts}
    \item[] Question: Does the paper discuss both potential positive societal impacts and negative societal impacts of the work performed?
    \item[] Answer: \answerNA{}
    \item[] Justification: This is foundational interpretability research on publicly available pretrained models. It releases no new model or dataset and has no direct path to deployment, so we do not foresee direct positive or negative societal impacts beyond a better understanding of how language models learn.

\item {\bf Safeguards}
    \item[] Question: Does the paper describe safeguards that have been put in place for responsible release of data or models that have a high risk for misuse (e.g., pre-trained language models, image generators, or scraped datasets)?
    \item[] Answer: \answerNA{}
    \item[] Justification: The paper does not release a new pretrained language model, image generator, scraped dataset, or other high-risk artifact. It uses existing public checkpoints and public linguistic datasets for analysis.

\item {\bf Licenses for existing assets}
    \item[] Question: Are the creators or original owners of assets (e.g., code, data, models), used in the paper, properly credited and are the license and terms of use explicitly mentioned and properly respected?
    \item[] Answer: \answerYes{}
    \item[] Justification: \Cref{app:compute-licenses} credits Pythia, WikiText-103, UD English EWT, UNER English EWT, STREUSLE, BLiMP, the subject--verb agreement data of \citet{linzen-etal-2016-assessing,goldberg2019assessing}, and the software we build on, and states their licenses. We use these assets for academic analysis and do not redistribute raw corpus text.

\item {\bf New assets}
    \item[] Question: Are new assets introduced in the paper well documented and is the documentation provided alongside the assets?
    \item[] Answer: \answerYes{}
    \item[] Justification: The paper does not release new datasets or models. The released code (\url{https://github.com/YpLarryWang/concept-energy}) includes a README describing the environment, data preparation, and how to run the main experimental pipeline, and is licensed under Apache-2.0.

\item {\bf Crowdsourcing and research with human subjects}
    \item[] Question: For crowdsourcing experiments and research with human subjects, does the paper include the full text of instructions given to participants and screenshots, if applicable, as well as details about compensation (if any)?
    \item[] Answer: \answerNA{}
    \item[] Justification: The research does not involve crowdsourcing, new annotation by paid workers, user studies, or experiments with human subjects.

\item {\bf Institutional review board (IRB) approvals or equivalent for research with human subjects}
    \item[] Question: Does the paper describe potential risks incurred by study participants, whether such risks were disclosed to the subjects, and whether Institutional Review Board (IRB) approvals (or an equivalent approval/review based on the requirements of your country or institution) were obtained?
    \item[] Answer: \answerNA{}
    \item[] Justification: The research does not involve human-subject experiments or newly collected participant data, so IRB or equivalent approval is not applicable.

\item {\bf Declaration of LLM usage}
    \item[] Question: Does the paper describe the usage of LLMs if it is an important, original, or non-standard component of the core methods in this research? Note that if the LLM is used only for writing, editing, or formatting purposes and does \emph{not} impact the core methodology, scientific rigor, or originality of the research, declaration is not required.
    \item[] Answer: \answerNA{}
    \item[] Justification: LLMs are the objects of study; no LLM is part of the method. AI assistants used for coding and writing are disclosed in the acknowledgments.

\end{enumerate}

\end{document}